%% file: main.tex
\documentclass[10pt,journal,letterpaper]{IEEEtran}
\usepackage{cite}
\usepackage{amsmath,amssymb}
\usepackage{graphicx}
\usepackage[hyphens]{url}
\usepackage{booktabs,array}
\usepackage[table]{xcolor}
\usepackage{mdframed}
\usepackage{algorithm}
\usepackage{algpseudocode}
\usepackage[caption=false,font=footnotesize]{subfig}
\usepackage{listings}
\usepackage{needspace}
\usepackage{float}
\usepackage{placeins}
\usepackage[nocheckfootnote]{flushend}
\usepackage[hidelinks]{hyperref}

\algrenewcommand{\algorithmiccomment}[1]{%
  \hfill\textcolor{gray}{\(\triangleright\) #1}%
}
\lstdefinestyle{amscontract}{
  basicstyle=\ttfamily\normalsize,
  columns=fullflexible,
  keepspaces=true,
  showstringspaces=false,
  breaklines=true,
  breakatwhitespace=false,
  breakindent=1em,
  postbreak=\mbox{\hspace{1em}},
  frame=single,
  framerule=0.3pt,
  framesep=3pt,
  aboveskip=6pt,
  belowskip=6pt,
  abovecaptionskip=4pt,
  belowcaptionskip=5pt,
  tabsize=2,
  numbers=none
}
\definecolor{amschemrow}{RGB}{235,243,250}
\DeclareRobustCommand{\revisionmark}[1]{#1}
\newmdenv[backgroundcolor=white,linewidth=0pt,
  leftmargin=0pt,rightmargin=0pt,innerleftmargin=3pt,innerrightmargin=3pt,
  innertopmargin=2pt,innerbottommargin=2pt,skipabove=0pt,skipbelow=0pt,
  nobreak=true]{revisionblock}
\newcommand{\method}{AMSchem}
\newcommand{\benchname}{\method{}-Bench}
\newcommand{\geomir}{SchematicIR}
\newcommand{\layoutsketch}{DrawingIntent}

\title{AMSchem: Automated Schematic Visualization of AMS Circuits via Verifiable Agentic Generation}
\input{authors}
\begin{document}
\raggedend
\maketitle

\begin{abstract}
\input{secs/0_abstract}
\end{abstract}
\begin{IEEEkeywords}
Analog and mixed-signal circuits, electronic design automation,
schematic generation, large language models, multimodal agents,
connectivity verification.
\end{IEEEkeywords}

\input{secs/1_intro}

\input{secs/2_related_work}
\input{secs/3_method}
\input{secs/4_experiments}
\FloatBarrier
\flushcolsend
\input{secs/6_conclusion}


\raggedcolsend
\bibliographystyle{IEEEtran}
\bibliography{refs}

\flushcolsend
\clearpage
\onecolumn
\appendices
\input{appendix/appendix}
\end{document}

%% file: authors.tex
\author{Yichen~Shi, Zhiyuan~Wang, Jianing~Cui, Yao~Lai,
Jinting~Zhang, Zhuofu~Tao, Hongyang~Wang,
Zhiping~Yu,~\IEEEmembership{Life~Fellow,~IEEE,}
Ting~Jung~Lin, and~Lei~He,~\IEEEmembership{Fellow,~IEEE}%
\thanks{\textit{Corresponding authors: Ting Jung Lin and Lei He.}}%
\thanks{Yichen Shi is with Shanghai Jiao Tong University, Shanghai, China,
and Eastern Institute of Technology, Ningbo, China.}%
\thanks{Zhiyuan Wang is with the University of Washington, USA.}%
\thanks{Jianing Cui is with the University of St Andrews, St Andrews, U.K.}%
\thanks{Yao Lai is with the University of Cambridge, Cambridge, U.K.}%
\thanks{Jinting Zhang is with Tianjin University, Tianjin, China.}%
\thanks{Zhuofu Tao is with the University of California, Los Angeles, CA, USA.}%
\thanks{Hongyang Wang is with the School of Computer Science and Technology,
Shijiazhuang Tiedao University, Shijiazhuang, China.}%
\thanks{Zhiping Yu is with Tsinghua University, Beijing, China.}%
\thanks{Ting Jung Lin and Lei He are with the Ningbo Institute of Digital Twin,
Eastern Institute of Technology, Ningbo, China
(e-mail: tlin@idt.eitech.edu.cn; lei.hexun@hotmail.com).}%
\\[0.4em]
{\normalfont\normalsize Project page:~\href{https://www.amschem.com/page}{\textcolor{blue!60!black}{\nolinkurl{www.amschem.com/page}}}}}

%% file: secs/0_abstract.tex
For decades, the netlist-to-schematic gap has remained a persistent challenge in design automation, especially for analog and mixed-signal (AMS) circuits. Netlists encode connectivity, not visual presentation, making their manual conversion into readable schematics time-consuming and dependent on expert conventions.
The same gap leaves LLM-based design tools without structured visual
feedback, forcing them to iterate on circuits they cannot see.
We present AMSchem, a training-free, verifier-guided agentic framework that automatically generates human-readable schematics from AMS netlists. AMSchem separates circuit-aware visual planning (DrawingIntent) from executable schematic geometry (SchematicIR), and applies localized visual feedback with bounded, revalidated repair patches. Deterministic validators enforce device preservation, pin-net consistency, and geometric correctness throughout.
For evaluation, we construct AMSchem-Bench comprising 64 circuits of 12 types. On the 40-circuit schematic subset, AMSchem with gpt-5.5 achieves 85.0\% validity (vs. 32.5\% direct generation). When integrated as a visual feedback module in LLM-based analog design, AMSchem resolves all 13 previously failed cases from AnalogCoder benchmark across five models, substantially improving the success rate.
Our dataset and benchmark are available at \url{https://www.amschem.com/benchmark}.

%% file: secs/1_intro.tex
\section{Introduction}

Readable and editable schematics are the primary human-facing interface in analog and mixed-signal (AMS) circuit design. Netlists specify devices, pins, parameters, and connectivity, but do not encode the visual organization that makes a circuit understandable. Hence, designers heavily rely on schematics in iterative design flows.

\begin{figure}[!t]
  \centering
  \includegraphics[width=\columnwidth]{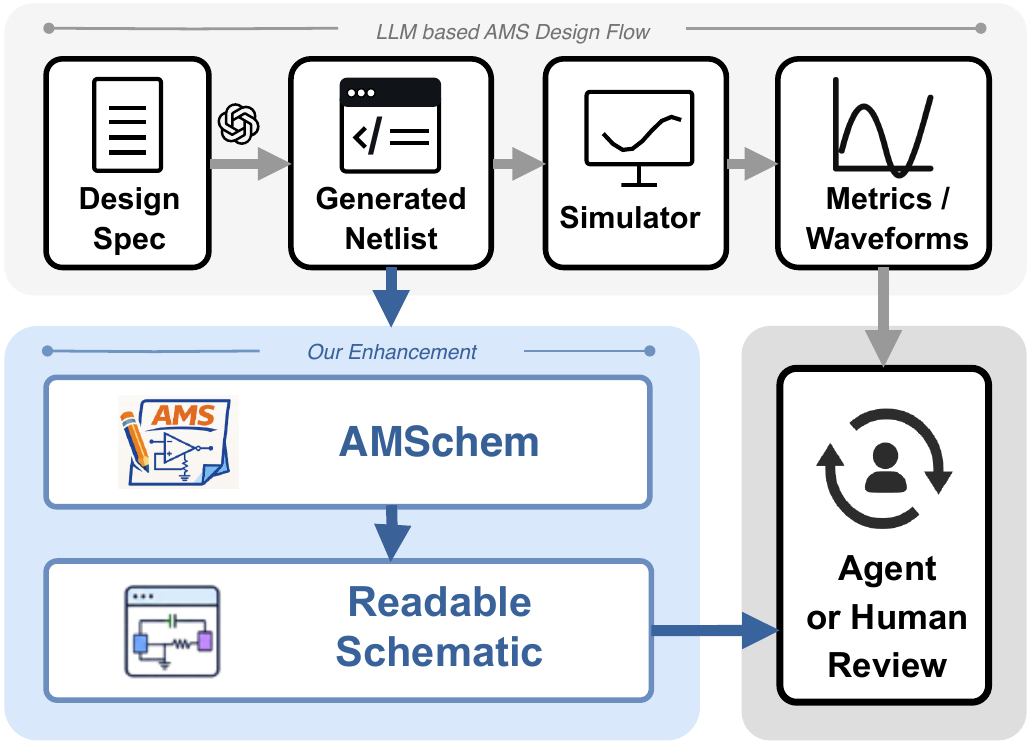}
  \caption{\method{} in LLM-driven AMS design flow: \method{} converts netlists into readable schematics, providing visual feedback for design review and optimization by agents or humans.}
  \label{fig:amschem_positioning}
\end{figure}

Meanwhile, recent large language model (LLM)-based systems have shown promise for topology generation \cite{chang2024lamagic,lai2024analogcoder,liu2024ladac, shen2025atelier, chen2024artisan, vijayaraghavan2025autocircuit} and device sizing \cite{yin2024adollm,liu2025llmsizing,yu2026autosizer}. However, these efforts remain largely organized around textual circuit representations and numerical performance metrics. As a result, though LLM-generated netlists can satisfy electrical checks, they still lack the schematic artifacts designers critically rely on, hindering further optimization and deployment. Fig.~\ref{fig:amschem_positioning} illustrates this gap in current LLM-driven design flows.

The netlist-to-schematic problem is a long-standing, widely anticipated yet unsolved challenge in electronic design automation (EDA). Despite decades of effort in graph drawing and automated layout \cite{sugiyama1981methods,gansner1993technique, hsu2022automatic, yang2023aem, du2026asgf4epd}, the objective remains difficult because connectivity alone underdetermines the visual realization of a circuit. Many drawings can preserve the same topology, but only a small subset follows designer conventions for functional grouping, signal direction, symmetry, port placement, and labeling. These circuit-dependent readability preferences must be reconciled with hard constraints on device preservation, pin-to-net consistency, and geometric legality. Consequently, generic graph-drawing objectives and handcrafted layout rules struggle to generalize across heterogeneous AMS circuits.

Recent efforts have explored LLM or multimodal LLM (MLLM)-based schematic construction \cite{matsuo2024schemato,liu2025eeschematic}. However, circuit knowledge does not directly translate into reliable schematic drawing. Unlike textual netlists, where a parser can validate syntax incrementally, schematic geometry admits no such intermediate verification. Although some methods use iterative visual refinement, preserving source-netlist connectivity remains challenging. A local error in symbol placement, terminal orientation, or wire routing can silently corrupt the entire schematic, while visual plausibility offers no guarantee of topological fidelity. Consequently, direct LLM generation remains brittle for producing schematics that are both readable and topologically correct.

This paper presents AMSchem, a training-free, verifier-guided agentic framework with a three-stage generation-and-refinement pipeline for the netlist-to-schematic task. 
The first stage matches drawing-relevant AMS building blocks and instantiates corresponding drawing references to construct DrawingIntent, a coordinate-free plan for circuit-aware organization. Second, it grounds the DrawingIntent into SchematicIR, an executable and machine-checkable representation of schematic geometry with source-netlist provenance.
Finally, AMSchem renders the candidate and applies localized visual feedback to identify readability issues. Deterministic validators are interleaved throughout the pipeline. Detected semantic, geometric, and visual issues are translated into bounded SchematicIR patches that are revalidated before acceptance, enabling local repair while preserving source connectivity.

The contributions are summarized as follows.
\begin{enumerate}
  \item AMSchem: a training-free, verifier-guided agentic framework for automatic AMS netlist-to-schematic visualization.
  \item Three-stage generation-and-refinement: (1) DrawingIntent as circuit-aware drawing plan, (2) SchematicIR as executable schematic geometry, and (3) localized visual feedback for structured repair.
  \item AMSchem-Bench: a comprehensive benchmark for AMS schematic visualization comprising 64 circuits of 12 circuit types and various complexity levels, with systematic metrics for schematic correctness and readability.
\end{enumerate}

On the 40-circuit schematic subset of AMSchem-Bench, AMSchem with gpt-5.5 achieves 85.0\% validity versus 32.5\% for direct generation, with overall gains in readability. When integrated as a visual feedback module in LLM-based analog design tasks in AnalogCoder benchmark~\cite{lai2024analogcoder}, AMSchem resolves all 13 previously failed cases across five models.

%% file: secs/2_related_work.tex
\section{Related Work}

\subsection{LLMs for AMS Circuit Design}
LLMs have recently been explored for AMS circuit design automation \cite{chang2024lamagic,lai2024analogcoder,yin2024adollm,liu2025llmsizing,yu2026autosizer,bujana2026selfcalibrating,pham2025genieasi}.
Prior work demonstrates LLMs' capability of encoding circuit-design priors and orchestrating simulators or analysis tools.
However, existing LLM-based methods typically produce text-based outputs, such as circuit netlists, scripts, device parameters, and simulation-based performance results.
Engineers, however, rely on readable schematics for design comprehension and debugging, as their visual form captures accumulated design experience and implicit constraints.
\method{} targets this missing netlist-to-schematic interface by converting a netlist into a structured, verifiable, and visual schematic state rather than stopping at text, code, or metric optimization.

\subsection{Automatic Schematic Visualization}

Prior to recent LLM-based approaches, \cite{hsu2022automatic} explored automatic analog schematic generation by combining circuit building-block classification with reinforcement learning; evaluation focuses on block compliance and wire crossings, but the approach requires extensive training and does not generalize across unseen topologies. Schemato \cite{matsuo2024schemato} directly generates LTSpice or CircuiTikz rendering code from netlists, entangling connectivity, geometry, and drawing syntax in a single output sequence without an intermediate editable representation. EEschematic \cite{liu2025eeschematic} employs an MLLM with visual in-context examples for schematic generation, but its evaluation is limited to a small handful of circuits and lacks systematic topological validity checking. Weave \cite{gulgonul2026weave} applies deterministic layered graph layout with round-trip netlist verification, yet treats the circuit as a generic graph without explicit modeling of functional grouping, symmetry, or signal-flow direction. 
Consequently, existing methods do not jointly provide circuit-aware planning, source-grounded executable representations, and verifier-guided local repair. AMSchem addresses this gap by separating DrawingIntent from SchematicIR and combining deterministic verification with bounded visual refinement. Prior evaluations also use disparate datasets, output formats, and metrics, hindering direct comparison and motivating AMSchem-Bench as a unified benchmark of schematic correctness and readability.

%% file: secs/3_method.tex
\section{Method}
\label{sec:method}

Fig.~\ref{fig:amschem_full_pipeline} illustrates the \method{} netlist-to-schematic flow for AMS circuits. \method{} follows a three-stage generation-and-refinement pipeline. The three refinement levels (L1--L3) share one rule: no revision is accepted without passing deterministic validation.

\subsection{DrawingIntent Generation and Refinement}
\label{app:bbl}

\method{} separates circuit-aware visual planning from concrete geometry
generation through an intermediate representation called \layoutsketch{}.
This stage proceeds in three steps: \emph{matching} circuit structures against predefined building block library, \emph{organization} of these
structures into a coordinate-free visual plan, and
\emph{constraint-driven refinement} to ensure plan consistency before
geometry grounding.

\begin{figure*}[!t]
  \centering
  \includegraphics[width=\textwidth]{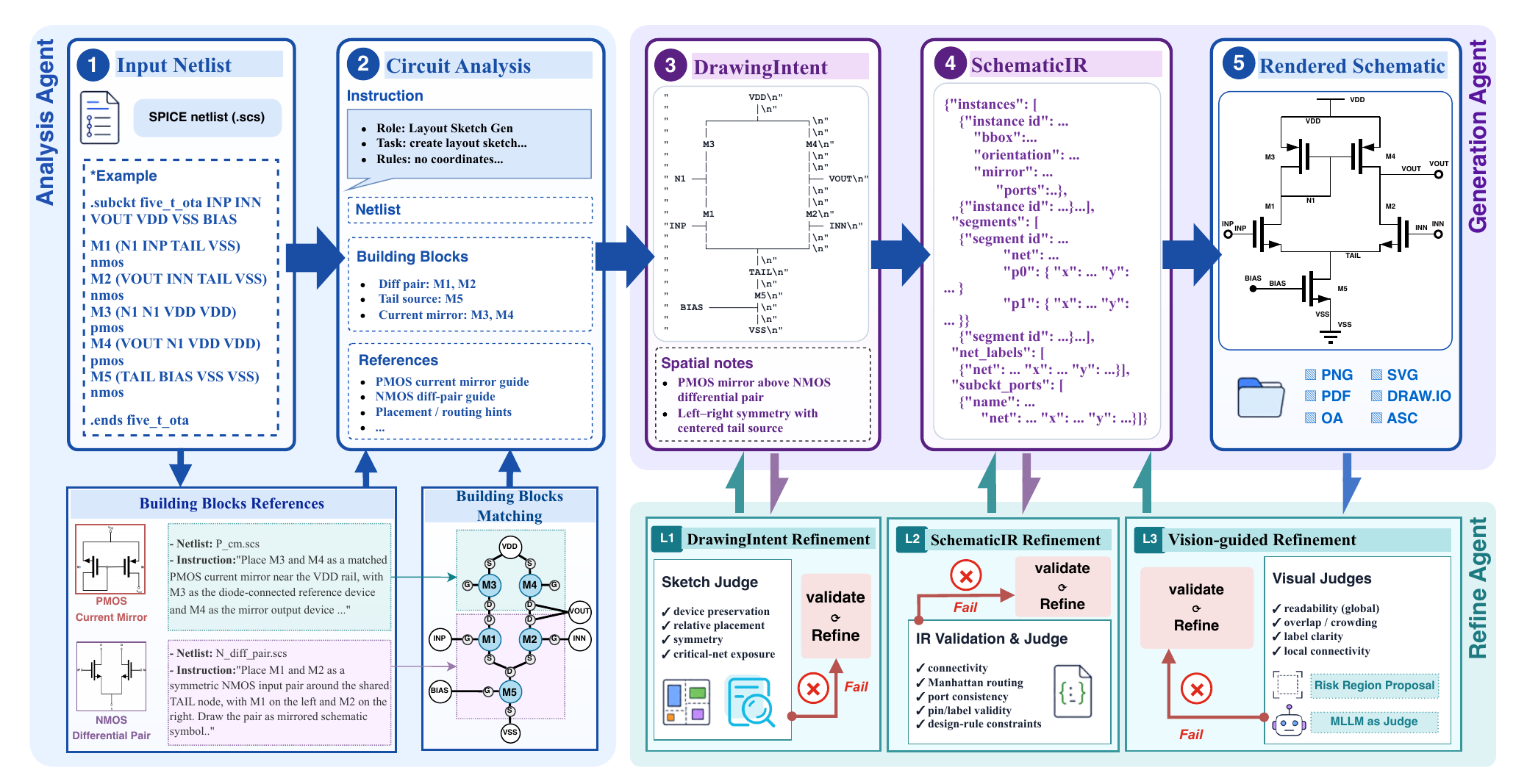}
  \caption{\method{} overview: (1) Stage 1: the input netlist is enriched with matched building blocks and drawing references to construct \textbf{\layoutsketch{}}, with constraint-driven refinement (L1) of device preservation, relative placement, and symmetry. (2) Stage 2: \textbf{\layoutsketch{}} is translated into executable \textbf{\geomir{}}, with structured validation and refinement of connectivity (L2), Manhattan routing, and design-rule compliance. (3) Stage 3: the rendered schematic undergoes bounded visual refinement (L3) that diagnoses readability, overlap, and local connectivity, producing the final output artifacts.}
  \label{fig:amschem_full_pipeline}
\end{figure*}

\paragraph{Building-Block Matching.}

The core data structure of this stage is a \emph{paired reference
library} $\mathcal{R}=\{R_k\}_{k=1}^{K}$, in which each reference
$R_k=(N_k^{\mathrm{ref}},\Pi_k^{\mathrm{ref}})$ couples a netlist
fragment with its preferred visual organization.  The unit of knowledge
is the \emph{pairing} itself: $N_k^{\mathrm{ref}}$ specifies the
topology and the functional roles of a common AMS building block
(e.g., two-transistor NMOS current mirror, two-branch PMOS cascode current mirror),
while $\Pi_k^{\mathrm{ref}}$ is an drawing instruction, capturing how 
that block is conventionally drawn---symmetry relations, device orientation, vertical ordering,
path separation, and the exposure of feedback connections.  Table~\ref{tab:bbl-library}
lists the 18 references currently in $\mathcal{R}$; the library is
extensible by adding further pairs without modifying the matching
machinery described below.

Formally, our building-block matching is \emph{type-constrained subgraph
matching over this paired library}.  The netlist $N$ is first parsed
into a typed circuit graph $G=(V,E,\tau)$, where $\tau$ assigns to
each vertex a device type (e.g., \textsf{nmos}, \textsf{pmos},
\textsf{res}) and to each incident edge a terminal role
(e.g., \textsf{gate}, \textsf{drain}, \textsf{source}, \textsf{bulk}).
The type function $\tau$ is what makes generic subgraph matching
circuit-aware: it rules out matches that are topologically isomorphic
but functionally different---for instance, a diode-connected device and
a common-source device can share the same bare topology, yet only the
latter satisfies the role constraints of a differential-pair input.

Each reference netlist $N_k^{\mathrm{ref}}$ is parsed into a typed
graph $G_k^{\mathrm{ref}}$ under the same type system, and matches are
computed by monomorphism search
\begin{equation}
\mathcal{B}=\mathcal{M}(G,\mathcal{R})
            =\bigcup_{k=1}^{K}\bigl\{(k,\phi):\phi\in
              \mathrm{VF3}(G,G_k^{\mathrm{ref}};\tau)\bigr\},
\label{eq:matching}
\end{equation}
where VF3~\cite{carletti2018vf3} is adopted as the monomorphism engine and
$\mathrm{VF3}(\cdot\,;\tau)$ denotes a search restricted to
type-compatible mappings.  Our design in this step lies in three
components built around the adopted engine: (i)~the type system above,
which encodes pin-role constraints absent from generic graph matching;
(ii)~symmetry reduction, which merges mappings that differ only by a
functional automorphism of the reference (e.g., swapping the two
branches of a current mirror), preventing an exponential blow-up of
equivalent matches; and (iii)~conflict pruning, which removes any match
whose device set is a strict subset of another match's, so that the
retained guidance is non-redundant.  A per-reference cap $L$ bounds
the number of matches kept for each $R_k$. 
\textbf{Algorithm~\ref{alg:bbl-algorithm}} summarizes the complete
procedure.

\begin{table}[H]
  \centering
  \caption{Building-block library used by \method{}.}
  \label{tab:bbl-library}
  \normalsize
  \renewcommand{\arraystretch}{1.05}
  \begin{tabular}{@{}r >{\raggedright\arraybackslash}p{0.86\columnwidth}@{}}
    \toprule
    \# & Building block \\
    \midrule
    1  & Two-transistor NMOS current mirror \\
    2  & Two-branch NMOS cascode current mirror \\
    3  & Two-transistor PMOS current mirror \\
    4  & Two-branch PMOS cascode current mirror \\
    5  & Single NMOS common-source device \\
    6  & Single PMOS common-source device \\
    7  & Two adjacent NMOS current-mirror pairs \\
    8  & Four-branch NMOS cascode mirror array \\
    9  & Two adjacent PMOS current-mirror pairs \\
    10 & Four-branch PMOS cascode mirror array \\
    11 & NMOS pair with a shared gate and source rail, and separate drains \\
    12 & PMOS pair with a shared gate and source rail, and separate drains \\
    13 & NMOS differential pair with a tail device \\
    14 & PMOS differential pair with an upper tail device \\
    15 & Two NMOS cascode devices sharing a gate bias \\
    16 & Two PMOS cascode devices sharing a gate bias \\
    17 & Two NMOS common-source devices tied to VSS \\
    18 & Two PMOS common-source devices tied to VDD \\
    \bottomrule
  \end{tabular}
\end{table}

\begin{algorithm}[!t]
\caption{\texorpdfstring{\revisionmark{Building-Block Matching and Drawing Guidance Generation}}{Building-Block Matching and Drawing Guidance Generation}}
\label{alg:bbl-selection}
\label{alg:bbl-algorithm}
\small
\begin{revisionblock}
\algrenewcommand{\algorithmicforall}{\textbf{for each}}
\begin{algorithmic}[1]
\Require Source netlist $N$, reference library $\mathcal{R}$
\Ensure Matched blocks $\mathcal{B}$, instantiated guidance $\mathcal{A}$
\State $G\gets\Call{TypedGraph}{N}$
\State $\mathcal{B}\gets\varnothing$; $\mathcal{A}\gets\varnothing$
\ForAll{$R_k\in\mathcal{R}$ in descending order of block size}
    \State $G_k\gets\Call{TypedGraph}{N_k^{\mathrm{ref}}}$
    \State $\mathcal{M}_k\gets\Call{VF3Monomorphism}{G_k,G}$ \Comment{\footnotesize Types/pin roles}
    \State $\mathcal{M}_k\gets\Call{SymmetryReduce}{\mathcal{M}_k}$ \Comment{\footnotesize Merge symmetries}
    \State $\mathcal{M}_k\gets\Call{Cap}{\mathcal{M}_k,L}$ \Comment{\footnotesize Per-reference cap}
    \State $\mathcal{B}\gets\mathcal{B}\cup\{(k,\phi)\mid\phi\in\mathcal{M}_k\}$
\EndFor
\State $\mathcal{B}\gets\{b\in\mathcal{B}\mid\nexists b'\in\mathcal{B}:V(b)\subsetneq V(b')\}$ \Comment{\footnotesize Remove subsets}
\ForAll{$(k,\phi)\in\mathcal{B}$}
    \State $\mathcal{A}\gets\mathcal{A}\cup\{\Pi_k^{\mathrm{ref}}[\phi]\}$ \Comment{\footnotesize Bind reference roles}
\EndFor
\State \Return $\mathcal{B},\mathcal{A}$
\end{algorithmic}
\end{revisionblock}
\end{algorithm}

Each retained match $b=(k,\phi)$ records the reference index $k$ and a
mapping $\phi$ from the functional roles of $N_k^{\mathrm{ref}}$ to
concrete devices and nets of $N$; $V(b)$ denotes the matched device
set.  Role binding instantiates the paired drawing instruction as
$\Pi_k^{\mathrm{ref}}[\phi]$, e.g., the abstract prior ``mirror the two
branches'' becomes ``place M1 and M2 mirror-symmetric about the
vertical axis.''  The resulting guidance set $\mathcal{A}$ is fused
with the source netlist into a circuit-specific generation context
\begin{equation}
C_D=\mathcal{C}(N,\mathcal{A}),
\label{eq:context}
\end{equation}
which preserves the original device and pin-to-net connectivity and
augments it with functional roles, structural relations, and
instantiated drawing principles.  All steps in this stage---parsing,
matching, symmetry reduction, pruning, and role binding---are
deterministic; no model inference is involved.

\paragraph{Visual Organization.}

Conditioned on the generation context $C_D$, the generation agent
produces an initial \emph{DrawingIntent} $D_0$: a coordinate-free,
circuit-aware plan that fixes \emph{how the circuit should be
organized in the schematic}.
For the OTA in Fig.~\ref{fig:amschem_full_pipeline}, for instance, such a plan
decides that the PMOS mirror pair sits above the NMOS differential
pair with the tail device centered on the symmetry axis, that inputs
enter from the left and the output leaves on the right---structural
commitments that can still be realized by many different concrete
layouts.

Formally, the DrawingIntent comprises two complementary components,
\begin{equation}
D_0=(P_0,\Pi_0)=F_D(C_D).
\label{eq:intent}
\end{equation}
The
\emph{Schematic Prototype} $P_0$ provides a coordinate-free structural
sketch specifying functional grouping, relative placement, signal-flow
direction, branch ordering, symmetry relations, and port distribution.
The \emph{Drawing Principles} $\Pi_0$ describe how the prototype should be
realized: preferred device orientations, labeling strategies, routing
preferences, treatment of global nets, and separation of signal, bias,
and feedback paths. The prototype captures \emph{what} to draw; the
principles govern \emph{how} to draw it.

\paragraph{Constraint-Driven Refinement.}

Refinement in AMSchem operates at three levels---intent,
geometry, and rendered image---denoted L1, L2, and L3 in
Fig.~\ref{fig:amschem_full_pipeline}.  This stage performs L1 refinement: all
repair happens on the coordinate-free plan, before any geometry
exists.

Before geometry generation, the L1 refinement agent validates the
\layoutsketch{} against three consistency criteria: coverage of the \revisionmark{matched}
building-block guidance, validity of referenced circuit objects, and
internal consistency of grouping, ordering, symmetry, and routing
preferences. Detected violations are represented as typed feedback
$\mathcal{E}_D$, and the affected intent fields are revised:
\begin{equation}
D=\mathrm{Refine}_D(D_0,\mathcal{E}_D).
\end{equation}

This intent-level refinement prevents incomplete or conflicting
high-level decisions from propagating into concrete symbols, coordinates,
and routed connections. The refined \layoutsketch{} $D=(P,\Pi)$ is then
passed to the \geomir{} stage for executable geometry grounding.

\subsection{SchematicIR Generation and Refinement}

Given the refined \layoutsketch{} $D$, \method{} constructs executable
\geomir{}---\textbf{the geometry-level ground of the pipeline: the
first representation in which the drawing plan becomes concrete
geometry, and, because every element is bound to the source netlist
through stable identifiers, the first artifact that is
machine-checkable.} 

\paragraph{Geometry Grounding.}

\begin{figure*}[!t]
  \centering
  \includegraphics[width=0.6\textwidth]{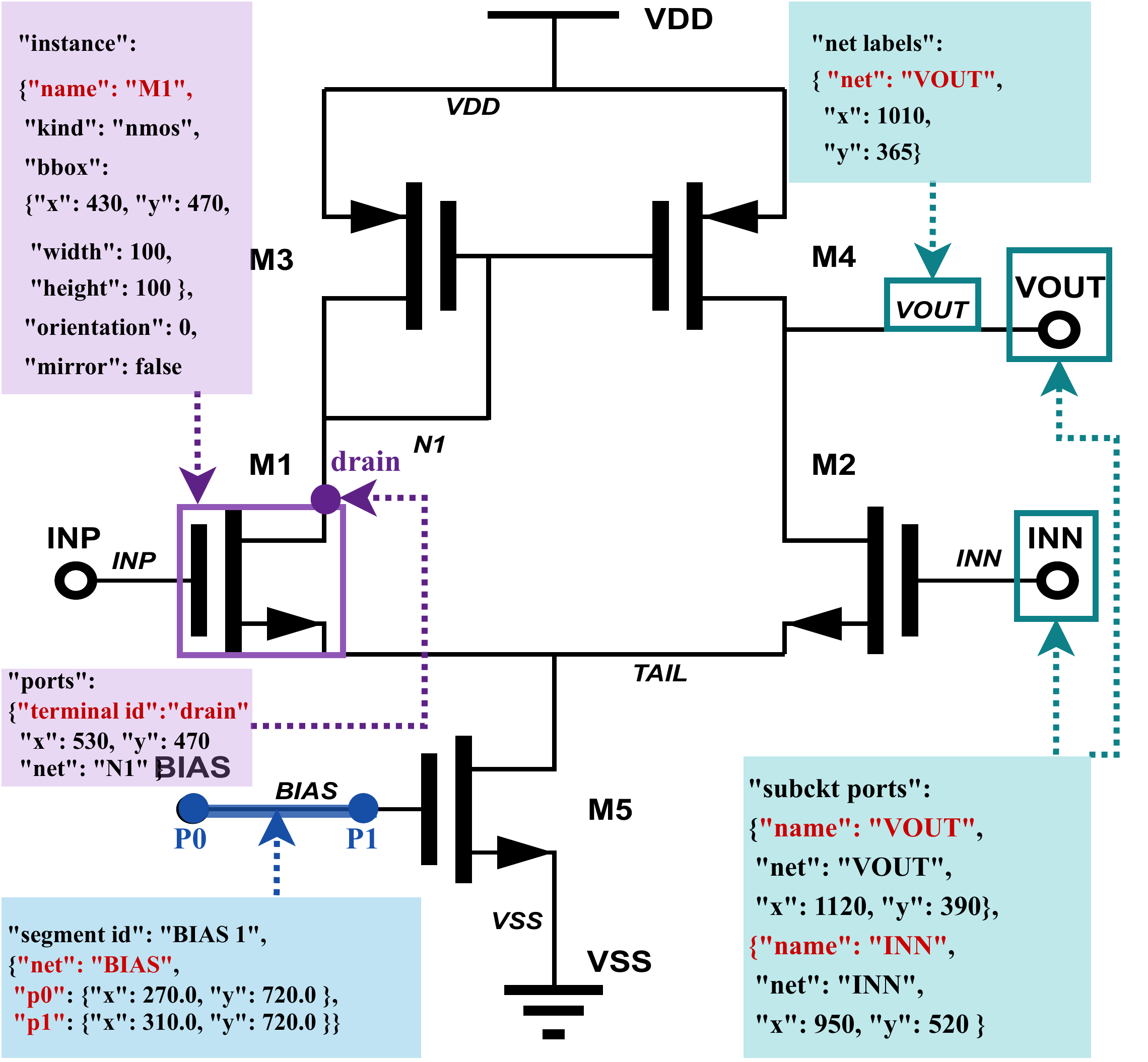}
  \caption{Object-level grounding in \geomir{}. Colored callouts link schematic elements to structured placement, orientation, geometry, and net bindings for deterministic validation and localized repair.}
  \label{fig:schematicir_representation}
\end{figure*}

The generation agent produces an initial \geomir{} $S_0$ from the source netlist and the refined \layoutsketch{}:
\begin{equation}
S_0=F_S(N,D).
\end{equation}
This transformation instantiates the structural prototype $P$ and drawing
principles $\Pi$ in $D$ as concrete symbols, coordinates, terminal anchors,
ports, labels, and routed wire segments, with junctions induced by
coincident endpoints of same-net segments. Source correspondence is
retained through stable instance, terminal and net identifiers. These bindings make the resulting schematic executable, machine-checkable, and locally repairable. As illustrated in Fig.~\ref{fig:schematicir_representation}, every visible
schematic element can therefore be traced to its \geomir{} object and
source-netlist entity.

Unlike rendering-code-level intermediates that entangle connectivity,
geometry, and drawing syntax in a single sequence~\cite{matsuo2024schemato},
\geomir{} keeps semantic bindings (identifiers, nets) separate from
visual realization (coordinates, segments), so that either side can be
repaired without disturbing the other.

\paragraph{Structured Validation.}

Deterministic validators check $S_0$ and represent all detected
violations as a unified issue set:
\begin{equation}
\mathcal{E}_S
=
\mathcal{E}_{\mathrm{net}}
\cup
\mathcal{E}_{\mathrm{short}}
\cup
\mathcal{E}_{\mathrm{overlap}}
\cup
\mathcal{E}_{\mathrm{rule}}.
\end{equation}
Here, $\mathcal{E}_{\mathrm{net}}$ captures inconsistencies between the normalized device--terminal--net relations induced from \geomir{} and those in the source netlist. $\mathcal{E}_{\mathrm{short}}$ records unintended electrical contacts between different nets; $\mathcal{E}_{\mathrm{short}}$
records unintended electrical contacts between different nets;
$\mathcal{E}_{\mathrm{overlap}}$ captures prohibited geometric
intersections among wires, symbols, and labels; and
$\mathcal{E}_{\mathrm{rule}}$ contains the remaining schema, symbol,
terminal, routing, and renderability violations. Each violation is localized to specific \geomir{} objects, enabling focused repair without regenerating the complete schematic.

\paragraph{Bounded Refinement.}

L2 refinement repairs the geometry. It consumes the issue set $\mathcal{E}_S$ and revises the
implicated \geomir{} objects together with their direct dependencies,
leaving the rest of the schematic untouched.

\begin{equation}
\widetilde{S}=\mathrm{Refine}_S(S_0,\mathcal{E}_S).
\end{equation}

Simple violations are handled through deterministic transformations, such as reconnecting a wire endpoint, separating an unintended contact, or rerouting a wire around a component. Complex issues trigger object-grounded repair requests, and the resulting complete \geomir{} candidate is revalidated after each update. A repaired candidate $\widetilde{S}$ is accepted as $S$ only if it resolves the reported issue, preserves source-netlist connectivity, and satisfies
\begin{equation}
\mathcal{E}_S(\widetilde{S})=\varnothing.
\label{eq:l2-accept}
\end{equation}
The accepted \geomir{} $S$ is then passed to the localized visual-feedback stage.

\subsection{Localized Visual Feedback and Refinement}

\method{} next turns from correctness to readability: it deliberately
renders the accepted \geomir{} $S$ into an image,
\begin{equation}
I=\mathrm{Render}(S),
\end{equation}
and inspects the rendering for the class of defects that structured
validation cannot see---visually ambiguous contacts, obscured
terminals, or dense connections that no deterministic rule expresses,
yet that decide whether a human or an agent can trace the circuit.
L3 refinement therefore operates on rendered pixels, in contrast to
L1 (plan level) and L2 (geometry level). 
The localized visual-feedback loop proceeds in three steps: proposing
circuit-aware risk regions where readability defects are likely,
diagnosing each cropped region with an object-grounded visual model,
and translating actionable feedback into bounded \geomir{} revisions
under the same revalidation gate as L2.

\paragraph{Risk-Region Proposal.}

The first step of the loop surveys the rendering as a whole before
narrowing scrutiny.  A whole-image
review of $I$ checks global readability---overall balance, spacing,
and label legibility---after which Risk Region Proposal (RRP) selects
the circuit-aware regions likely to contain localized readability
issues:
\begin{equation}
\mathcal{R}=\mathrm{RRP}(I,S)
=\mathcal{R}_{\mathrm{BBL}}\cup\mathcal{R}_{\mathrm{patch}}
\cup\mathcal{R}_{\mathrm{highdeg}}.
\label{eq:rrp}
\end{equation}
Here, $\mathcal{R}_{\mathrm{BBL}}$ prioritizes matched building blocks
whose visual organization matters for circuit understanding,
$\mathcal{R}_{\mathrm{patch}}$ covers previously modified objects and
their local neighborhoods, and $\mathcal{R}_{\mathrm{highdeg}}$ targets
high-degree nets and their connected components.

\begin{figure*}[!t]
  \centering
  \includegraphics[width=0.72\textwidth]{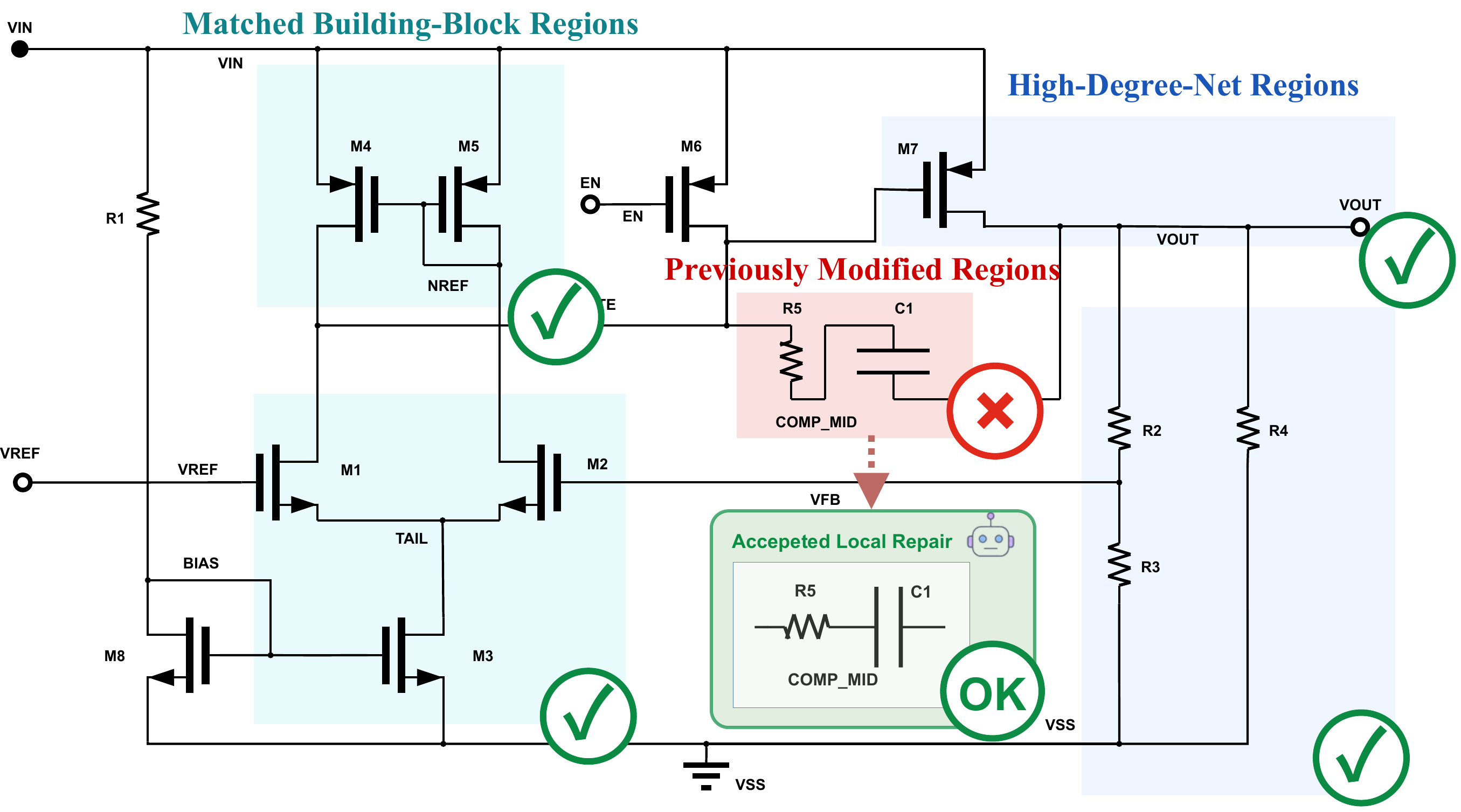}
  \caption{L3 refinement: localized visual feedback over circuit-aware
risk regions.  Regions proposed by RRP (a) are cropped and diagnosed
by an object-grounded visual model (b); actionable findings guide a
bounded \geomir{} revision (c) that is revalidated and rerendered.}
  \label{fig:divide_refine_visual_feedback}
\end{figure*}

\paragraph{Object-Grounded Visual Diagnosis.}

The second step examines each proposed region with a visual model.
For every selected region $R_i\in\mathcal{R}$, \method{} extracts the
corresponding crop,
\begin{equation}
I_i=\mathrm{Crop}(I,R_i).
\end{equation}
and presents the visual model $F_V$ with the circuit semantics behind
the pixels: the associated \geomir{} objects $O_i$, the local netlist
context $N_i^{\mathrm{loc}}$, and the relevant \layoutsketch{} fields
$D_i$:
\begin{equation}
u_i=F_V\bigl(I_i,\,O_i,\,N_i^{\mathrm{loc}},\,D_i\bigr).
\label{eq:diagnosis}
\end{equation}

The grounding is what makes the output actionable. Both the
rendered pixels and the intended connectivity are in context. Hence, the
model can judge, for example, whether a visually ambiguous junction
actually joins the nets the netlist requires, and report the affected
objects, the local readability issue, and a concrete \geomir{}
correction in each feedback item $u_i$.

\paragraph{Bounded Visual Refinement.}

The third step closes the loop by turning the diagnosed feedback into
geometry, under the same bounded discipline as L2. 
The visual refinement agent translates the aggregated feedback
 $u_i$ into a revised \geomir{} candidate,
\begin{equation}
\widetilde{S}^{(i)}=\mathrm{Refine}_V\left(S^{(i-1)},u_i\right).
\end{equation}
The revision may reroute wires, adjust labels or junctions, replace long
wires with net labels, or reposition local groups. As in L2, the
revision is bounded in both senses: the agent restricts changes to
the objects identified by $u_i$ and their dependencies, and the
complete candidate is accepted and then rerendered for the next
whole-image review only if it preserves connectivity and passes
structured validation (Eq.~\ref{eq:l2-accept}).

Fig.~\ref{fig:divide_refine_visual_feedback} illustrates this region-guided diagnosis and repair process. Because $\mathcal{R}_{\mathrm{patch}}$
re-examines previously modified regions, each accepted repair feeds
back into the next round of review, and the loop terminates when a
rendering raises no further actionable feedback.

%% file: secs/4_experiments.tex
\begin{figure*}[!t]
  \centering

  \subfloat[Source composition and circuit-scale statistics.]{\label{fig:benchmark-source-composition}%
\begin{minipage}[c]{0.7\textwidth}
    \centering
    \includegraphics[width=\linewidth]{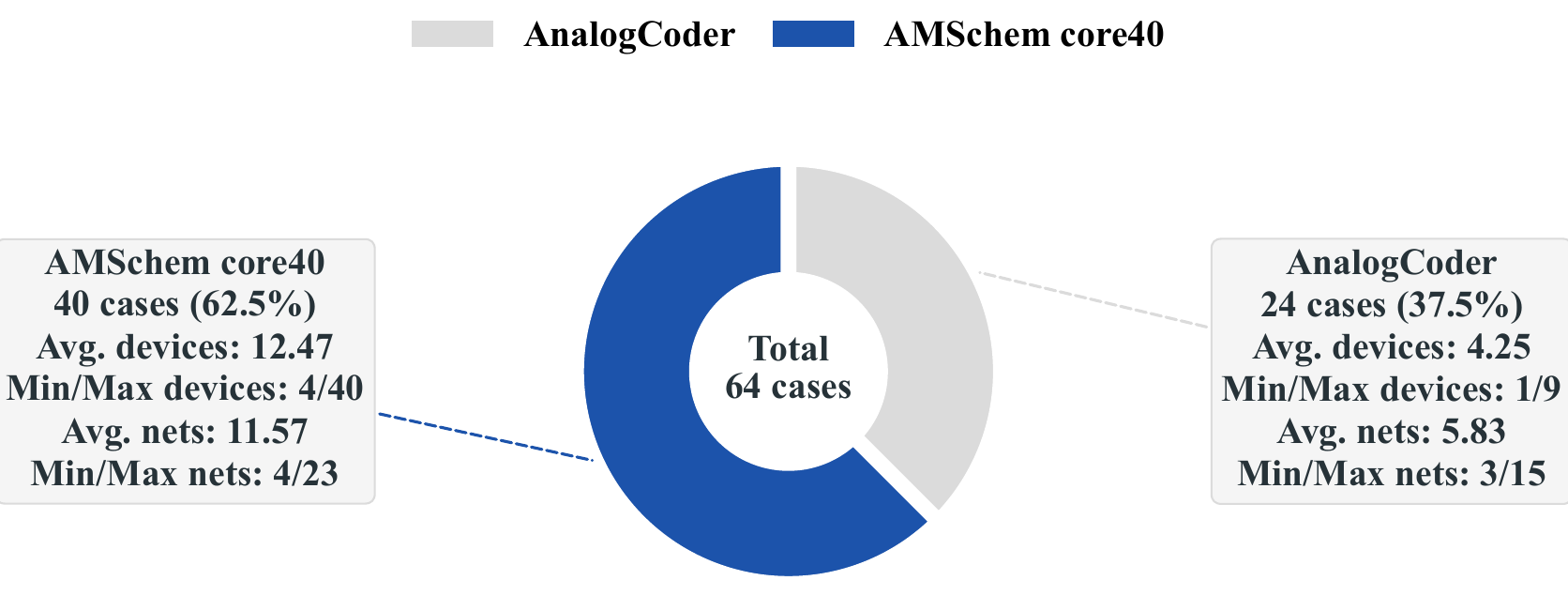}

  \end{minipage}}

  \vspace{1mm}

  \subfloat[Statistics of circuit type and difficulty tier.]{\label{fig:benchmark-circuit-distribution}%
\begin{minipage}[c]{0.7\textwidth}
    \centering
    \includegraphics[width=\linewidth]{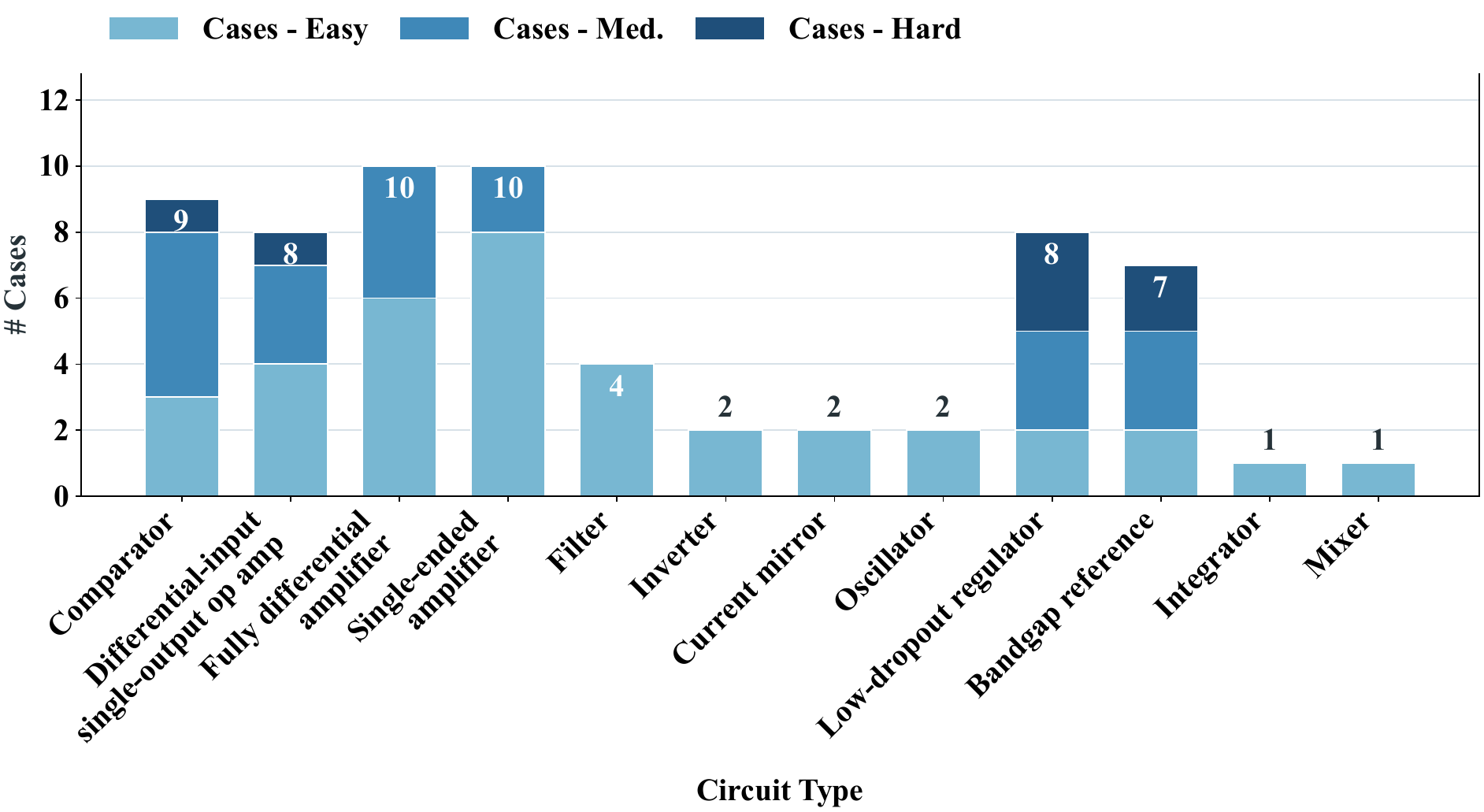}

  \end{minipage}}

  \caption{\benchname{} statistics.}
  \label{fig:benchmark-composition}
\end{figure*}

\section{Experiments}

\subsection{\benchname{}}

Prior AMS schematic visualization methods are evaluated under inconsistent protocols: some verify only render success or syntax compliance~\cite{matsuo2024schemato}, others use readability metrics based on small-scale subjective assessments~\cite{liu2025eeschematic}, and none jointly assess topological fidelity and readability with standardized metrics and datasets. Consequently, systematic comparison across methods remains impossible. To address this, we introduce \benchname{}, a benchmark suite of 64 circuits covering 12 circuit types and comprising two subsets used in separate experiments: 40 circuits from AMSnet 2.0~\cite{shi2025amsnet20largeams} for netlist-to-schematic evaluation and 24 circuits from AnalogCoder~\cite{lai2024analogcoder} for downstream LLM-aided circuit-design evaluation.

The summarized statistics in Fig.~\ref{fig:benchmark-composition} show that the two sources differ substantially in circuit scale. To characterize circuit complexity consistently, we define three source-independent difficulty tiers based on device count: Easy ($\leq 9$ devices), Medium (10--19), and Hard ($\geq 20$). 

\input{tabs/1_direct_ir_main_results}

\subsection{Evaluation Metrics}
\label{sec:metrics}

We evaluate generated \geomir{} schematic for validity and readability. Readability is assessed using reference-based metrics for routing quality and component organization, complemented by subjective mean opinion scores (MOS).

\noindent\textbf{Validity.}
An output is valid if it preserves source-netlist connectivity, conforms
to the \geomir{} schema, uses legal symbol and terminal references,
obeys deterministic wiring constraints, renders successfully, and
avoids wires crossing unrelated device bodies.
We additionally report $R_{\mathrm{overlap}}$, 
the mean number of wire crossings through
unrelated device bodies, as a continuous measure of the same
geometric-legality requirement; it is averaged over all rendered
outputs, so it also reflects how nearly invalid outputs miss the
threshold.

Let $v_n\in\{0,1\}$ denote case $n$'s validity. Over $N$ cases, the validity rate is
\begin{equation}
P_{\mathrm{valid}}
=
\frac{1}{N}
\sum_{n=1}^{N} v_n.
\end{equation}

\noindent\textbf{Routing Quality.}
For a valid output $S$ and its expert reference $S^\star$, let $C(S)$ and $B(S)$ denote the numbers of non-connective wire crossings and Manhattan-route bends, respectively. We define
\begin{equation}
\begin{aligned}
S_{\mathrm{cross}}
&=
1-\frac{|C(S)-C(S^\star)|}
{\max\{1,C(S),C(S^\star)\}}, \\
S_{\mathrm{bend}}
&=
1-\frac{|B(S)-B(S^\star)|}
{\max\{1,B(S),B(S^\star)\}}.
\end{aligned}
\end{equation}

\noindent\textbf{Component Organization.}
Components are matched by their source-netlist instance identifiers. Let $\mathcal{I}_m$ be the set of matched instances and $\mathcal{P}$ the set of comparable instance pairs. For each $(i,j)\in\mathcal{P}$, $r_S(i,j)$ denotes their relative spatial relation (left, right, above, or below), while $o_S(i)$ denotes the orientation of instance $i$, including rotation and mirroring. We define
\begin{equation}
\begin{array}{@{}l@{\;}c@{\;}l@{}}
\displaystyle S_{\mathrm{rel}}
& = &
\displaystyle
\frac{1}{|\mathcal{P}|}
\sum_{(i,j)\in\mathcal{P}}
\mathbf{1}\!\left[r_S(i,j)=r_{S^\star}(i,j)\right],
\\[0.8ex]
\displaystyle S_{\mathrm{orient}}
& = &
\displaystyle
\frac{1}{|\mathcal{I}_m|}
\sum_{i\in\mathcal{I}_m}
\mathbf{1}\!\left[o_S(i)=o_{S^\star}(i)\right].
\end{array}
\end{equation}
All four readability scores lie in $[0,1]$, with higher values indicating closer agreement with the reference. At the dataset level, invalid outputs are assigned zero for all four metrics before averaging. 

Because multiple layouts may represent the same topology equally well, we additionally report MOS from a blinded study in which 15 engineers independently rated each generated schematic on a five-point scale.
\subsection{Experimental Setup}

All methods share the same input netlists, renderer, and evaluation metrics.
\emph{Direct-IR} generates the complete \geomir{} from the netlist in one pass. \emph{Direct-IR + Few-shot} adds $K$ in-context examples to the same
one-pass prompt.  \emph{\method{}} runs the full pipeline of
Sec.~\ref{sec:method}.  Each strategy is instantiated with every
backbone MLLM (gpt-5.5, gemini-3.1-pro, claude-opus-4-8). The result is summarized in Table~\ref{tab:direct-ir-main-results}.

We also include Weave~\cite{gulgonul2026weave}, a deterministic layered-graph
visualizer with round-trip netlist verification, as the representative
of non-LLM visualization. EESchematic produced no valid outputs on our cases and is reported qualitatively in the Appendix, while Schemato is excluded because no public implementation is available. For the downstream LLM-aided circuit-design experiment, we follow the original AnalogCoder protocol and simulation setting.

\subsection{Main Results}

\begin{figure*}[!t]
\centering
\begin{minipage}[t]{0.34\textwidth}
\vspace{0pt}
\centering
\subfloat[Ground truth]{\label{fig:visual-case-gt}%
\includegraphics[width=\linewidth]{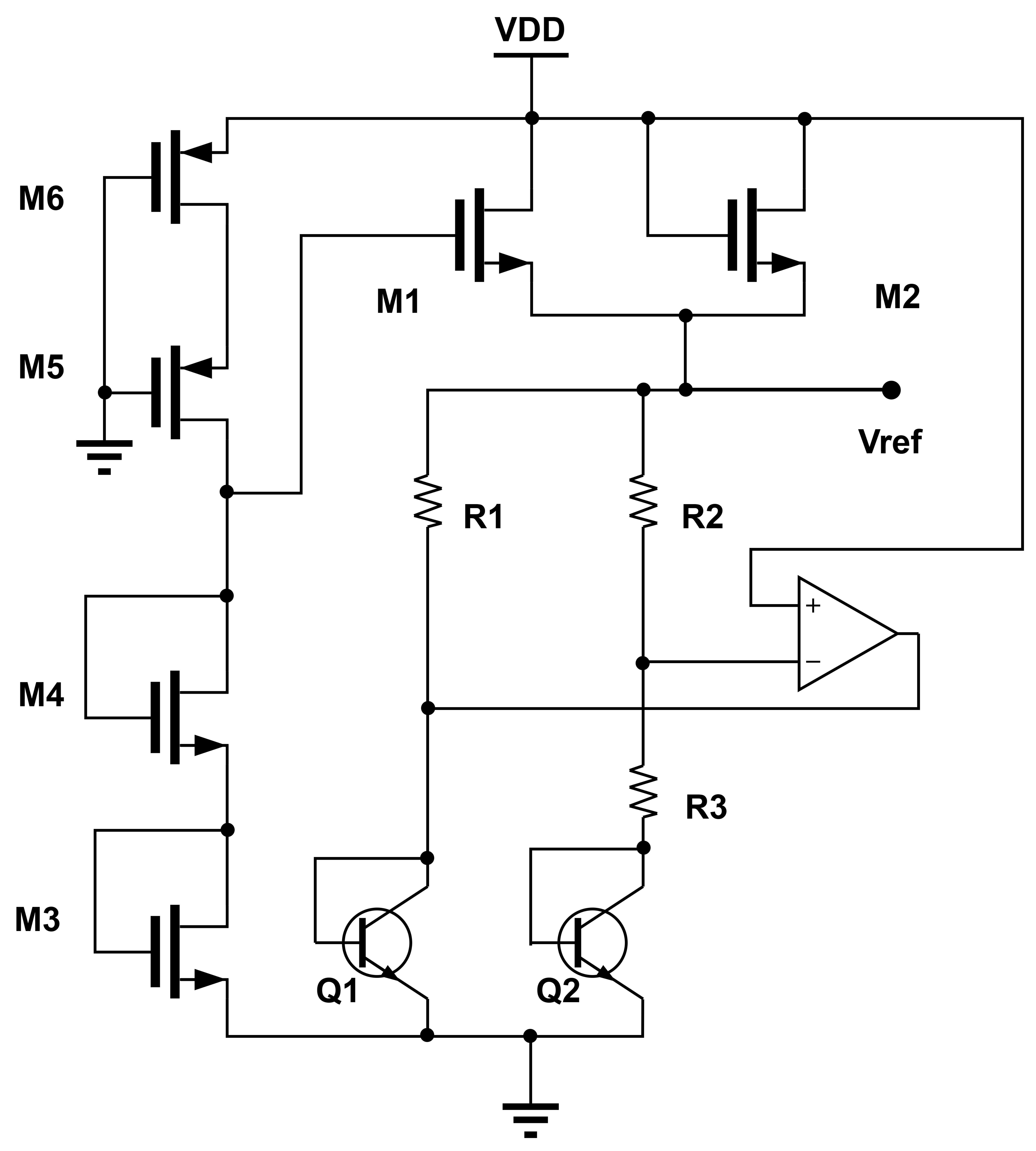}}
\end{minipage}\hfill
\begin{minipage}[t]{0.42\textwidth}
\vspace{0pt}
\centering
\subfloat[\method{}]{\label{fig:visual-case-amschem}%
\includegraphics[width=\linewidth]{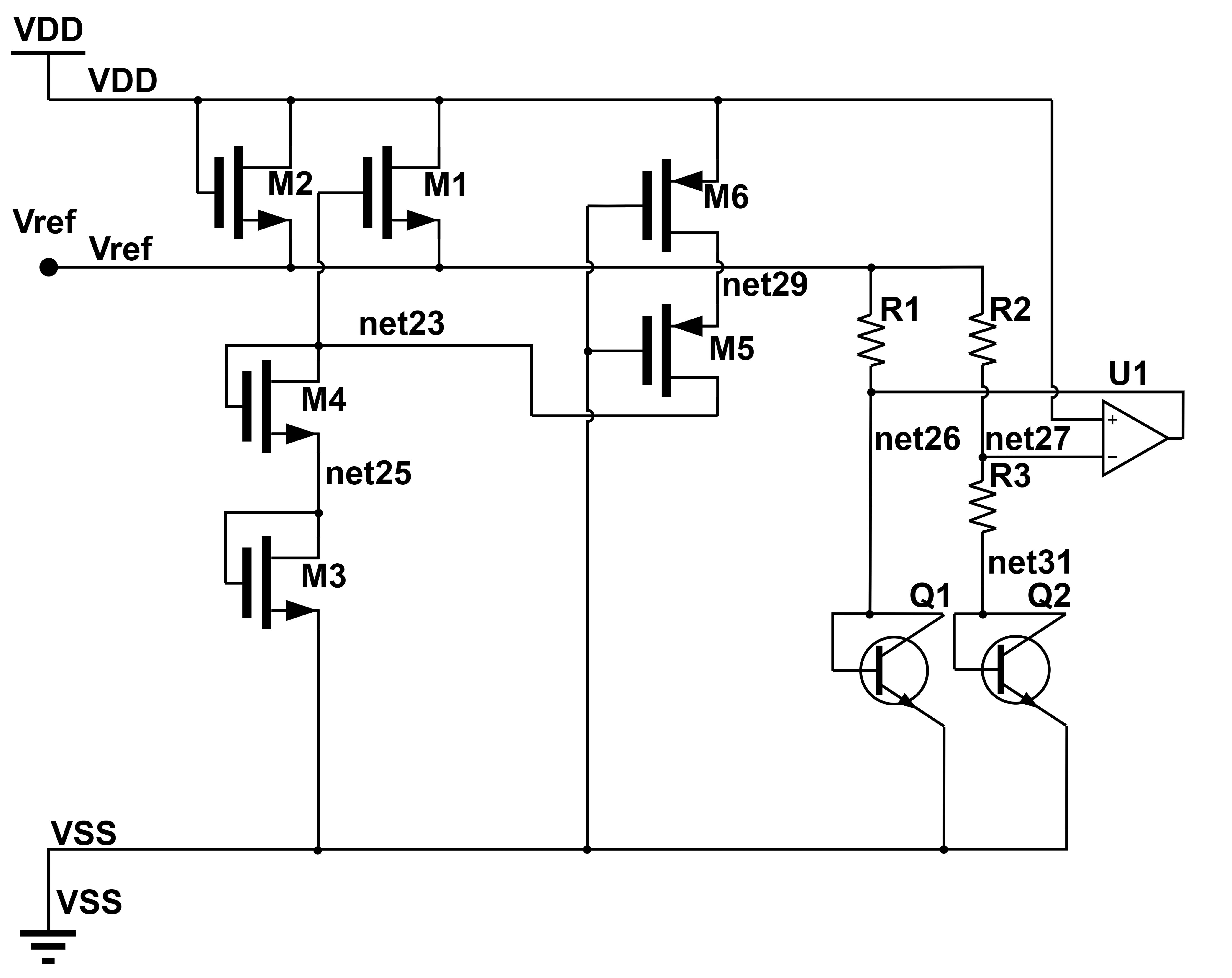}}
\end{minipage}\hfill
\begin{minipage}[t]{0.22\textwidth}
\vspace{0pt}
\centering
\subfloat[Weave]{\label{fig:visual-case-weave}%
\includegraphics[width=\linewidth,height=0.27\textheight,keepaspectratio]{figs/Visual_case_Weave.pdf}}
\end{minipage}
\caption{Schematic comparison on a bandgap reference circuit from
AMSchem-Bench: (a) Ground truth; (b) \method{}; (c)
Weave~\cite{gulgonul2026weave}.}
\label{fig:visual_case}
\end{figure*}

\noindent\textbf{Validity.}
Weave first marks the limit of the deterministic approach: its layered
layout attains moderate validity on Easy and Medium circuits (69.2\%
and 60.0\%) but collapses on Hard circuits (28.6\%),
whereas \method{} maintains 92.3\%, 95.0\%, and 42.9\% on the same
tiers with gpt-5.5 (Table~\ref{tab:direct-ir-main-results}).  
Among generative methods,
\method{} improves validity over both \emph{Direct-IR} baselines for
every evaluated MLLM and difficulty tier, with particularly large
gains on Medium and Hard circuits: for gpt-5.5, from the best
baseline results of 10.0\% and 0.0\% to 95.0\% and 42.9\%,
respectively; for gemini-3.1-pro, from 35.0\% and 14.3\% to 90.0\%
and 71.4\%.

In contrast, \emph{Direct-IR + Few-shot} does not reliably help: for
gpt-5.5 it lowers validity on the Easy tier from 84.6\% to 69.2\%,
and analogous degradations appear for the other backbones.  Few-shot
examples bias the model toward reproducing the example layouts, but
they do not encode the constraints that couple device organization to
routing; on circuits whose correct organization differs from the
examples, the referenced patterns may violate connectivity or
geometry rules.

\noindent\textbf{Readability.}
Readability improves together with validity, though with smaller and
more variable margins.  On Hard circuits with gemini-3.1-pro, \method{}
raises $S_{\mathrm{cross}}$ from 0.0094 to 0.3616 and
$S_{\mathrm{rel}}$ from 0.0348 to 0.5000 relative to
\emph{Direct-IR}, with similar gains in the other reference-based
metrics (Table~\ref{tab:direct-ir-main-results}).  Qualitatively,
Fig.~\ref{fig:visual_case} shows that \method{} preserves the compact
two-dimensional organization of the ground-truth schematic, whereas
Weave \cite{gulgonul2026weave} arranges related elements along a long vertical path. Human judgment agrees: in the blinded MOS study, \method{}
scores 3.075, 4.128, and 3.432 under the three backbones, all well
above Weave's 1.150.  Read together, the reference-based metrics,
the qualitative comparison, and the blinded ratings show that
\method{} is the only configuration that improves validity and
readability at once: deterministic layout alone (Weave) remains
correct but unreadable, and unconstrained generation alone
(\emph{Direct-IR}) remains unreliable.

\subsection{Ablation Study}

\input{tabs/2_stage_wise_ablation}

\noindent\textbf{Contribution of Pipeline Stages.}
Table~\ref{tab:stage-wise-ablation} evaluates the incremental contribution of each pipeline stage. The three switches correspond to the
refinement levels of Sec.~\ref{sec:method}: DI enables L1 (DrawingIntent),
IR enables L2 (\geomir{} validation and repair), and VF enables L3
(visual feedback).  Recall from Sec.~\ref{sec:metrics} that
$R_{\mathrm{overlap}}$ is averaged over all rendered outputs, so it
reflects both the legality of valid outputs and how nearly invalid
outputs miss the geometric-legality threshold.  Enabling only DI
leaves validity unchanged at 32.5\%, consistent with the role of L1,
which acts on the plan before any geometry exists. Its effect is
purely organizational, nearly doubling the wire-crossing score and
reducing $R_{\mathrm{overlap}}$ from 5.550 to 1.763.  Adding IR yields
the largest validity gain, raising it to 80.0\% while improving all
readability metrics, showing that L2's validation gate is the pipeline's primary correctness mechanism.  VF further raises validity to 85.0\% and
reduces $R_{\mathrm{overlap}}$ from 0.900 to 0.475, a 47.2\%
reduction, with the wire-bending score remaining comparable.  The
three stages are therefore complementary: L1 improves organization, 
L2 resolves the majority of validity
failures, and L3 removes residual failures together with localized
readability defects.

\noindent\textbf{Stage-wise Token Consumption.}
Fig.~\ref{fig:token-consumption} complements the stage-wise ablation with the cost profile of the three stages. The reconstructed mean total is 153.3k tokens per case, with \layoutsketch{}, \geomir{}, and visual feedback accounting for 25.9\%, 24.2\%, and 49.9\%, respectively. Input tokens, including textual context and image inputs, constitute 88.3\% of the total; stored visible responses account for the remaining 11.7\%. Visual feedback is therefore the most token-intensive stage. Its contribution extends beyond the additional five percentage points in validity to a substantial reduction in device--wire overlaps, showing that its benefits should be assessed in terms of both validity and readability. The input-heavy consumption profile suggests that reducing redundant context and selectively allocating visual inspection are promising directions for improving token efficiency.
\input{figs/token_consumption}

\input{tabs/3_drawingintent_ablation}
\noindent\textbf{\layoutsketch{} Ablation.}

We next ablate the components of Stage 1 within the L2--L3-enabled
configuration (rows 3--4 of Table~\ref{tab:stage-wise-ablation}).
Table~\ref{tab:drawingintent-ablation} shows that cumulatively
enabling building-block matching, references, and intent refinement
increases validity from 72.5\% to 85.0\%
. Both $S_{\mathrm{bend}}$ and
$S_{\mathrm{rel}}$ improve monotonically, while the small initial decline in
$S_{\mathrm{cross}}$ is recovered after references are introduced.
The full configuration achieves the best overall performance, indicating that the three components contribute in a cumulative fashion.

\noindent\textbf{Risk Region Proposal Analysis.}
We conduct a blinded expert evaluation of RRP on 40 cases, with AMS engineers
annotating schematic defects independently of the generated proposals.
Table~\ref{tab:risk-region-proposal-analysis} reveals distinct operating
points for the three proposal sources. 
$\mathcal{R}_{\mathrm{highdeg}}$ provides
broad coverage, accounting for 66.4\% of proposals and achieving the highest
recall (55.3\%); $\mathcal{R}_{\mathrm{BBL}}$ strikes a
more balanced precision--recall trade-off.
$\mathcal{R}_{\mathrm{patch}}$ contributes 8.6\%
of proposals with recall of 8.1\%, because its role is not broad defect
discovery but the precise re-examination of previously repaired
regions, where its precision is the highest (76.0\%).  The three
sources are complementary in function rather than in coverage.

\input{tabs/4_risk_region_proposal}

\noindent\textbf{Discussion.}
Taken together, the ablations delineate a division of labor among the
three refinement levels.  L1 (\layoutsketch{}) shapes organization
without affecting validity, as shown by the unchanged validity and
doubled wire-crossing score when only DI is enabled
(Table~\ref{tab:stage-wise-ablation}); its internal components contribute
cumulatively (Table~\ref{tab:drawingintent-ablation}).  L2 (\geomir{} validation
and repair) is the dominant source of validity, raising it from
32.5\% to 80.0\%.  L3 (visual feedback) then converts residual
failures and removes localized readability defects that no rule
expresses---at the highest token cost among the stages. 
The three-level design is
the minimal configuration that improves validity and readability at
once.

\subsection{\method{} for LLM-Based Analog Design}

To evaluate schematic feedback in automated analog circuit design, we integrate \method{} into the AnalogCoder refinement loop. For cases unresolved by the original loop, \method{} produces a schematic whose connectivity is verified against the input netlist. A visual critic examines the circuit structure, biasing relationships, and signal paths, then returns diagnostic feedback to the original design model for netlist revision. Each unresolved case is given at most $T=5$ refinement rounds; all 13 are resolved within $n=4$ rounds. Each revised design is evaluated using the original simulator and functional checker. Within the five-model, 120-evaluation scope shown in Fig.~\ref{fig:downstream-recovery}, the original AnalogCoder loop solves 107 cases without \method{} assistance, and \method{}-assisted refinement recovers the remaining 13, bringing the total to 120. All recovered designs pass independent re-evaluation.
\input{tabs/6_analogcoder_recovery}

\input{figs/gilbert_mixer_schematic_repair_figure}

\Needspace{3\baselineskip}
Fig.~\ref{fig:gilbert-mixer-schematic-repair} illustrates how schematic feedback translates into a concrete topology repair for a Gilbert-cell mixer generated by MiniMax-M3 (T19). The task requires difference- and sum-frequency products under 1.0-kHz RF and 1.2-kHz LO excitation. In the failed design, devices controlled by the same LO phase connect both RF branches to the same output node, rather than forming the required cross-coupled switching structure. This connectivity is consistent with the observed LO-dominated output. The \method{} schematic exposes the connections between the switching quad and the RF branches, allowing the visual critic to localize the error to specific devices and output nets. Following this feedback, the design model reconnects the drain of M4 from \texttt{Voutn} to \texttt{Voutp} and that of M6 from \texttt{Voutp} to \texttt{Voutn}, restoring the cross-coupled paths to the differential outputs. This example shows how a netlist-faithful schematic can support localized diagnosis of functional errors in the netlist itself.
\input{figs/gilbert_mixer_repair_figure}

\Needspace{3\baselineskip}
Fig.~\ref{fig:gilbert-mixer-repair} shows the corresponding functional change, with panels (a, c) before repair and (b, d) after repair. Spectra computed from uniformly resampled 5--20-ms output waveforms show that the failed circuit is dominated by the 1.2-kHz LO component, while its 200-Hz difference and 2.2-kHz sum components are only 0.021 and 0.013~mV. After repair, these components reach 55.47 and 53.36~mV, respectively, both exceeding the 1-mV threshold, while LO feedthrough is suppressed. The original functional checker also reports PASS for the repaired design. The plotted responses follow the original evaluator's automatic LO bias-selection procedure. In additional comparisons at the same LO common-mode voltage, the original circuit fails to meet the mixing requirement, whereas the repaired circuit satisfies it. Together, the two figures trace schematic-assisted design from a localized connectivity diagnosis through netlist revision to simulation-based confirmation of the target function.

%% file: tabs/1_direct_ir_main_results.tex
\begin{table*}[!t]
\centering
\caption{Performance across benchmark difficulty tiers. E, M, and H denote Easy, Medium, and Hard, respectively. MOS is pooled across tiers. Shaded rows indicate AMSchem.}
\label{tab:direct-ir-main-results}
\normalsize
\setlength{\tabcolsep}{1.5pt}
\renewcommand{\arraystretch}{1.04}
\begin{tabular}{@{}c*{16}{c}@{}}
\toprule
Method
& \multicolumn{3}{c}{Valid (\%) $\uparrow$}
& \multicolumn{3}{c}{$S_{\mathrm{cross}}\uparrow$}
& \multicolumn{3}{c}{$S_{\mathrm{bend}}\uparrow$}
& \multicolumn{3}{c}{$S_{\mathrm{rel}}\uparrow$}
& \multicolumn{3}{c}{$S_{\mathrm{orient}}\uparrow$}
& \multicolumn{1}{c}{MOS $\uparrow$} \\
\cmidrule(lr){2-4}
\cmidrule(lr){5-7}
\cmidrule(lr){8-10}
\cmidrule(lr){11-13}
\cmidrule(lr){14-16}
\cmidrule(lr){17-17}
& E & M & H
& E & M & H
& E & M & H
& E & M & H
& E & M & H
& \\
\midrule
Weave~\cite{gulgonul2026weave}
& 69.2 & 60.0 & 28.6
& 0.3333 & 0.3007 & 0.1905
& 0.2200 & 0.1248 & 0.0347
& 0.2374 & 0.1741 & 0.0611
& 0.3370 & 0.2237 & 0.0986
& 1.150 \\
\midrule
\multicolumn{17}{@{}l}{\textcolor{gray}{\textit{gpt-5.5}}} \\
Direct-IR
& 84.6 & 10.0 & 0.0
& 0.2051 & 0.0800 & 0.0000
& 0.3968 & 0.0597 & 0.0000
& 0.5067 & 0.0667 & 0.0000
& 0.6316 & 0.0700 & 0.0000
& -- \\
\begin{tabular}[c]{@{}c@{}}Direct-IR\\+ Few-shot\end{tabular}
& 69.2 & 5.0 & 0.0
& 0.2308 & 0.0200 & 0.0000
& 0.3475 & 0.0087 & 0.0000
& 0.4106 & 0.0282 & 0.0000
& 0.4974 & 0.0273 & 0.0000
& -- \\
\rowcolor{amschemrow}\textbf{\method{}}
& \textbf{92.3} & \textbf{95.0} & \textbf{42.9}
& \textbf{0.4308} & \textbf{0.3732} & \textbf{0.1111}
& \textbf{0.3721} & \textbf{0.3476} & \textbf{0.1911}
& \textbf{0.6557} & \textbf{0.6112} & \textbf{0.3031}
& \textbf{0.7496} & \textbf{0.6557} & \textbf{0.3070}
& 3.075 \\
\midrule
\multicolumn{17}{@{}l}{\textcolor{gray}{\textit{gemini-3.1-pro}}} \\
Direct-IR
& 61.5 & 30.0 & 14.3
& 0.1538 & 0.1875 & 0.0094
& 0.4016 & 0.1876 & 0.0248
& 0.4553 & 0.2186 & 0.0348
& 0.4872 & 0.2207 & 0.0929
& -- \\
\begin{tabular}[c]{@{}c@{}}Direct-IR\\+ Few-shot\end{tabular}
& 69.2 & 35.0 & 0.0
& 0.3333 & 0.2306 & 0.0000
& 0.4073 & 0.1651 & 0.0000
& 0.4453 & 0.2517 & 0.0000
& 0.5171 & 0.2536 & 0.0000
& -- \\
\rowcolor{amschemrow}\textbf{\method{}}
& \textbf{100.0} & \textbf{90.0} & \textbf{71.4}
& \textbf{0.3654} & \textbf{0.4414} & \textbf{0.3616}
& \textbf{0.6129} & \textbf{0.4658} & \textbf{0.2689}
& \textbf{0.7162} & \textbf{0.6152} & \textbf{0.5000}
& \textbf{0.8521} & \textbf{0.6582} & \textbf{0.5025}
& \textbf{4.128} \\
\midrule
\multicolumn{17}{@{}l}{\textcolor{gray}{\textit{claude-opus-4-8}}} \\
Direct-IR
& 53.8 & 5.0 & 0.0
& 0.2308 & 0.0500 & 0.0000
& 0.3210 & 0.0368 & 0.0000
& 0.3154 & 0.0389 & 0.0000
& 0.4205 & 0.0200 & 0.0000
& -- \\
\begin{tabular}[c]{@{}c@{}}Direct-IR\\+ Few-shot\end{tabular}
& 53.8 & 15.0 & 0.0
& 0.1538 & 0.0392 & 0.0000
& 0.3068 & 0.0831 & 0.0000
& 0.3725 & 0.1142 & 0.0000
& 0.4538 & 0.1093 & 0.0000
& -- \\
\rowcolor{amschemrow}\textbf{\method{}}
& \textbf{76.9} & \textbf{30.0} & \textbf{14.3}
& \textbf{0.2949} & \textbf{0.1268} & \textbf{0.0435}
& \textbf{0.3586} & \textbf{0.1269} & \textbf{0.0476}
& \textbf{0.4545} & \textbf{0.1723} & \textbf{0.0444}
& \textbf{0.5825} & \textbf{0.2138} & \textbf{0.1143}
& 3.432 \\
\bottomrule
\end{tabular}
\end{table*}

%% file: tabs/2_stage_wise_ablation.tex
\begin{table*}[t]
\centering
\caption{Incremental analysis of \method{}. DI, IR, and VF
denote \layoutsketch{}, \geomir{} validation and refinement, and visual feedback,
respectively.}
\label{tab:stage-wise-ablation}

\begingroup
\normalsize
\setlength{\tabcolsep}{8.095pt}
\renewcommand{\arraystretch}{1.12}

\definecolor{StageHeader}{RGB}{211,226,243}
\definecolor{StageOnBg}{RGB}{229,239,249}
\definecolor{StageOnFg}{RGB}{31,78,121}
\definecolor{StageOffBg}{RGB}{248,248,248}
\definecolor{StageOffFg}{RGB}{145,145,145}

\newcommand{\stagehead}[1]{%
  \cellcolor{StageHeader}\textbf{#1}}
\newcommand{\stageon}{%
  \cellcolor{StageOnBg}%
  \textcolor{StageOnFg}{\ensuremath{\checkmark}}}
\newcommand{\stageoff}{%
  \cellcolor{StageOffBg}%
  \textcolor{StageOffFg}{\ensuremath{\times}}}

\normalsize
\begin{tabular}{@{}ccc@{\hspace{7pt}}cccccc@{}}
\toprule
\stagehead{DI}
& \stagehead{IR}
& \stagehead{VF}
& Valid (\%) $\uparrow$
& $S_{\mathrm{cross}}\uparrow$
& $S_{\mathrm{bend}}\uparrow$
& $S_{\mathrm{rel}}\uparrow$
& $R_{\mathrm{overlap}}\downarrow$
& $S_{\mathrm{orient}}\uparrow$ \\
\midrule
\stageoff
& \stageoff
& \stageoff
& 32.5
& 0.1067
& 0.1588
& 0.1980
& 5.550
& 0.2403 \\

\stageon
& \stageoff
& \stageoff
& 32.5
& 0.2083
& 0.1674
& 0.2533
& 1.763
& 0.2488 \\

\stageon
& \stageon
& \stageoff
& 80.0
& 0.3344
& \textbf{0.3339}
& 0.5305
& 0.900
& 0.5916 \\

\stageon
& \stageon
& \stageon
& \textbf{85.0}
& \textbf{0.3461}
& 0.3282
& \textbf{0.5717}
& \textbf{0.475}
& \textbf{0.6252} \\
\bottomrule
\end{tabular}
\endgroup
\end{table*}

%% file: figs/token_consumption.tex
\begin{figure}[!t]
\centering
\begin{minipage}[c][0.23\textheight][c]{\linewidth}
\centering
\includegraphics[width=\columnwidth]{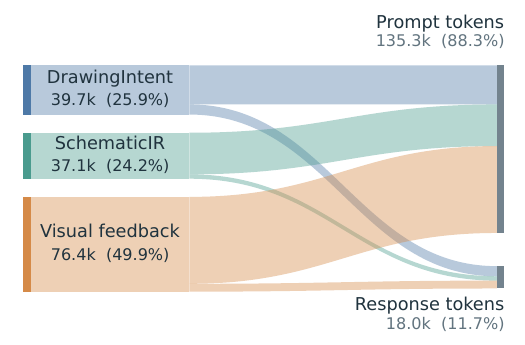}
\end{minipage}
\caption{Mean per-case token consumption by pipeline stage and token type.}
\label{fig:token-consumption}
\end{figure}

%% file: tabs/3_drawingintent_ablation.tex
\begin{table*}[!t]
\centering
\caption{Component ablation of \layoutsketch{}.}
\label{tab:drawingintent-ablation}

\begingroup
\setlength{\tabcolsep}{6pt}
\renewcommand{\arraystretch}{1.12}

\definecolor{StageHeader}{RGB}{211,226,243}
\definecolor{StageOnBg}{RGB}{229,239,249}
\definecolor{StageOnFg}{RGB}{31,78,121}
\definecolor{StageOffBg}{RGB}{248,248,248}
\definecolor{StageOffFg}{RGB}{145,145,145}

\newcommand{\stagehead}[1]{%
  \cellcolor{StageHeader}\textbf{#1}%
}
\newcommand{\stageon}{%
  \cellcolor{StageOnBg}%
  \textcolor{StageOnFg}{\ensuremath{\checkmark}}%
}
\newcommand{\stageoff}{%
  \cellcolor{StageOffBg}%
  \textcolor{StageOffFg}{\ensuremath{\times}}%
}

\normalsize
\begin{tabular}{@{}ccc@{\hspace{5pt}}ccccc@{}}
\toprule
\stagehead{BBL Match}
& \stagehead{BBL Ref.}
& \stagehead{Refine}
& Valid (\%) $\uparrow$
& $S_{\mathrm{cross}}\uparrow$
& $S_{\mathrm{bend}}\uparrow$
& $S_{\mathrm{rel}}\uparrow$
& $S_{\mathrm{orient}}\uparrow$ \\
\midrule

\stageoff
& \stageoff
& \stageoff
& 72.5
& 0.3212
& 0.2924
& 0.4787
& 0.5103 \\

\stageon
& \stageoff
& \stageoff
& 75.0
& 0.3141
& 0.3008
& 0.5099
& 0.5350 \\

\stageon
& \stageon
& \stageoff
& 77.5
& 0.3436
& 0.3030
& 0.5259
& 0.5721 \\

\stageon
& \stageon
& \stageon
& \textbf{85.0}
& \textbf{0.3461}
& \textbf{0.3282}
& \textbf{0.5717}
& \textbf{0.6252} \\

\bottomrule
\end{tabular}

\endgroup
\end{table*}

%% file: tabs/4_risk_region_proposal.tex
\definecolor{rrpHighDegree}{RGB}{26,86,183} 
\definecolor{rrpBBL}{RGB}{15,130,138}       
\definecolor{rrpModified}{RGB}{204,0,0}     

\newcommand{\rrpbox}[1]{%
  \raisebox{0.10em}{\textcolor{#1}{\rule{0.48em}{0.48em}}}%
}

\begin{table}[!htbp]
\centering

\caption{Comparison of RRP region types. Proposal share denotes
the fraction of proposals contributed by each type.}
\label{tab:risk-region-proposal-analysis}

\setlength{\tabcolsep}{3pt}
\renewcommand{\arraystretch}{1.12}

\normalsize
\begin{tabular}{@{}lcccc@{}}
\toprule

\shortstack{Risk\\region}
& \shortstack{Precision\\(\%) $\uparrow$}
& \shortstack{Recall\\(\%) $\uparrow$}
& \shortstack{F1 score\\(\%) $\uparrow$}
& \shortstack{Proposal\\share (\%)}
\\

\midrule

\rrpbox{rrpBBL}\hspace{0.45em}
$\mathcal{R}_{\mathrm{BBL}}$
& 69.9
& 21.7
& 33.1
& 25.0
\\

\rrpbox{rrpModified}\hspace{0.45em}
$\mathcal{R}_{\mathrm{patch}}$
& \textbf{76.0}
& 8.1
& 14.6
& 8.6
\\

\rrpbox{rrpHighDegree}\hspace{0.45em}
$\mathcal{R}_{\mathrm{highdeg}}$
& 67.0
& \textbf{55.3}
& \textbf{60.6}
& 66.4
\\

\bottomrule
\end{tabular}

\end{table}

%% file: tabs/6_analogcoder_recovery.tex
\begin{figure}[t]
\centering
\begin{minipage}[c][0.23\textheight][c]{\linewidth}
\centering
\includegraphics[width=0.9\columnwidth]{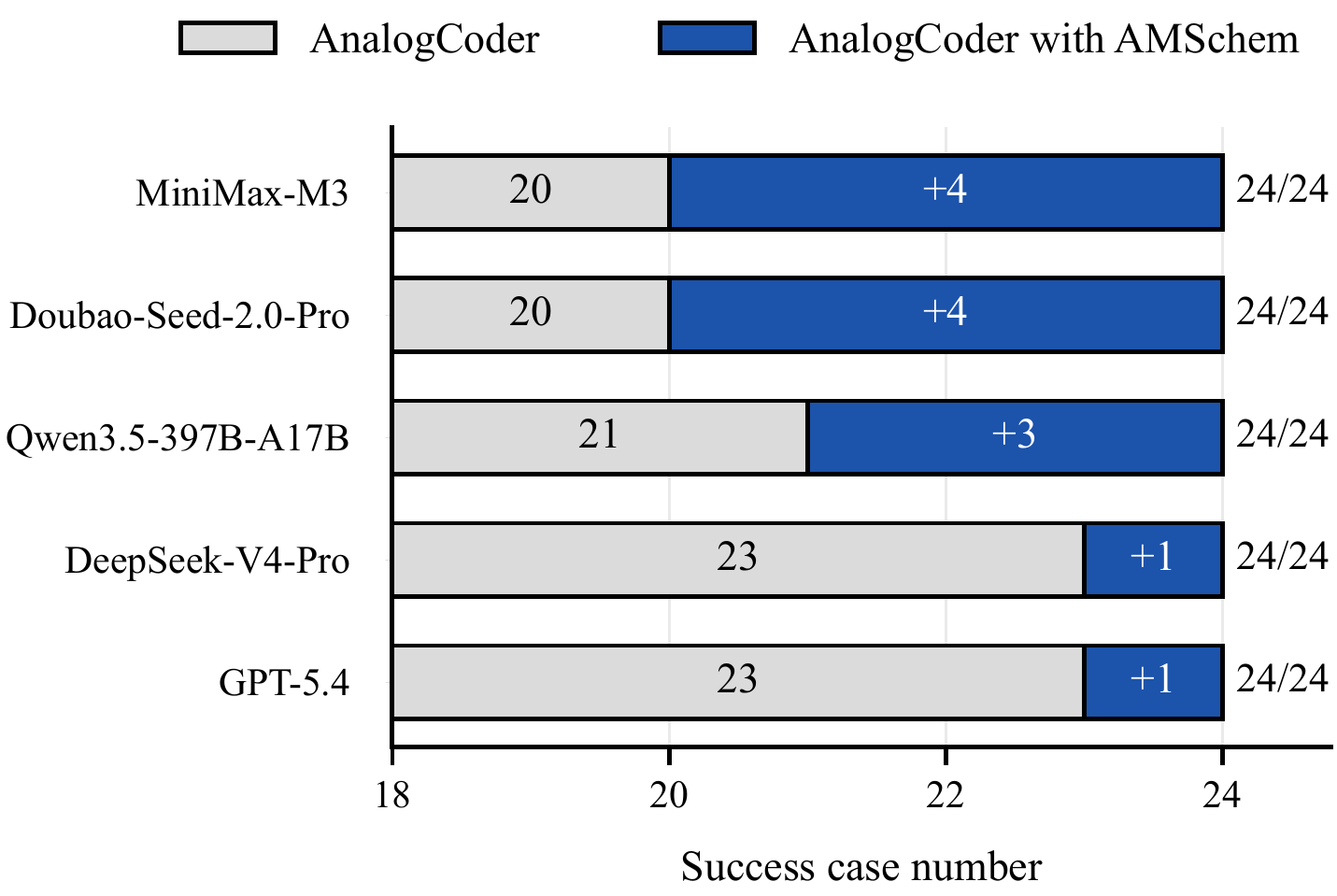}
\end{minipage}
\caption{Effect of \method{}-aided visual feedback in the AnalogCoder refinement loop across five models.}
\label{fig:downstream-recovery}
\end{figure}

%% file: figs/gilbert_mixer_schematic_repair_figure.tex
\begin{figure*}[!t]
\centering
\includegraphics[width=\textwidth]{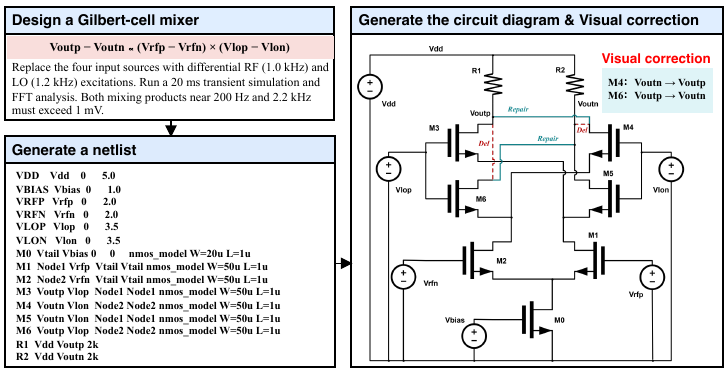}
\caption{Schematic-guided repair of a Gilbert-cell mixer generated by MiniMax-M3 (T19). The task and failed netlist are shown on the left. The schematic on the right identifies the incorrect output connections of the LO switching quad: red dashed segments mark removed connections, and teal segments show the drain reconnections of M4 and M6.}
\label{fig:gilbert-mixer-schematic-repair}
\end{figure*}

%% file: figs/gilbert_mixer_repair_figure.tex
\begin{figure*}[!t]
\centering
\includegraphics[width=0.69\textwidth]{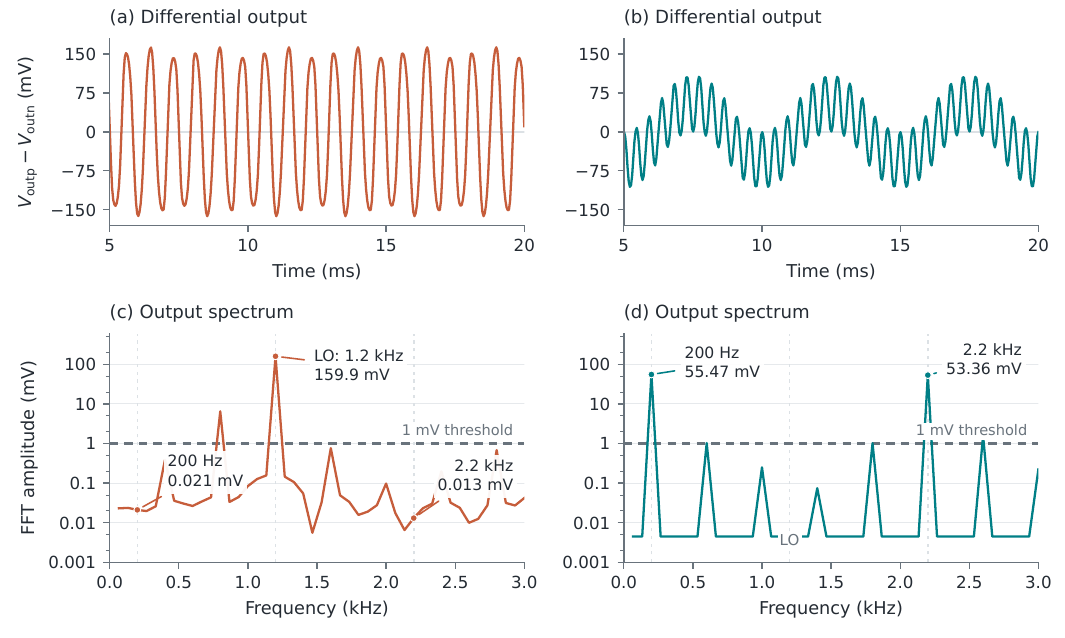}
\caption{Gilbert-cell mixer response before (a,c) and after (b,d)
schematic-guided repair.  (a,b) Differential output waveforms;
(c,d) FFT spectra of uniformly resampled 5--20-ms segments
(24~kSa/s sampling, $N_{\mathrm{FFT}}=360$, rectangular window).  Excitation:
1.0-kHz RF, 1.2-kHz LO.  The plotted traces follow the original
evaluator's automatic LO bias-selection procedure. Additional
comparisons at the same LO common-mode voltage show that the
original circuit fails the mixing requirement and the repaired
circuit satisfies it.  Dashed line: 1-mV product threshold.}
\label{fig:gilbert-mixer-repair}
\end{figure*}

%% file: secs/6_conclusion.tex
\section{Conclusion}

The netlist-to-schematic gap has outlived decades of graph-drawing
research because connectivity alone underdetermines a readable
drawing.  \method{} closes this gap with a principle that we expect to
outlast any single model generation: \emph{circuit-aware planning
proposes, and deterministic validation disposes}.  By separating
\layoutsketch{} (what to draw) from \geomir{} (verifiable geometry)
and gating every revision---at the plan, geometry, and pixel
levels---behind revalidation, \method{} converts an unreliable
generative task into a verifiable one.  The principle is model-agnostic:
it lifts three different backbone MLLMs from 32.5\% to 85.0\% validity
with consistent readability gains. Once embedded in an LLM
design loop, its schematic feedback recovers all 13 cases that a
state-of-the-art text-only loop left unsolved.

We regard the second finding as the more consequential one.  AMS design is not only about electrical behavior. How a circuit is
organized in the schematic carries design knowledge, and a design agent that
cannot see its own circuit is blind to exactly the class of
errors---miswired switches, broken symmetry, misplaced feedback---that
engineers can spot instantly.  \method{} turns the schematic from a
passive deliverable into a machine-checkable, agent-consumable
feedback interface inside the design loop.  Extending this interface
to hierarchical schematics, and coupling visual critique with
simulation evidence, is our immediate next step.

%% file: appendix/appendix.tex
\section{MOS Evaluation Protocol and Platform}
\label{app:mos}

\begin{figure}[H]
  \centering

  \subfloat[Landing page of the blinded MOS evaluation platform.]{\label{fig:mos-platform-login}%
\begin{minipage}[t]{0.48\textwidth}
    \centering
    \includegraphics[
      width=\linewidth,
      height=0.32\textheight,
      keepaspectratio
    ]{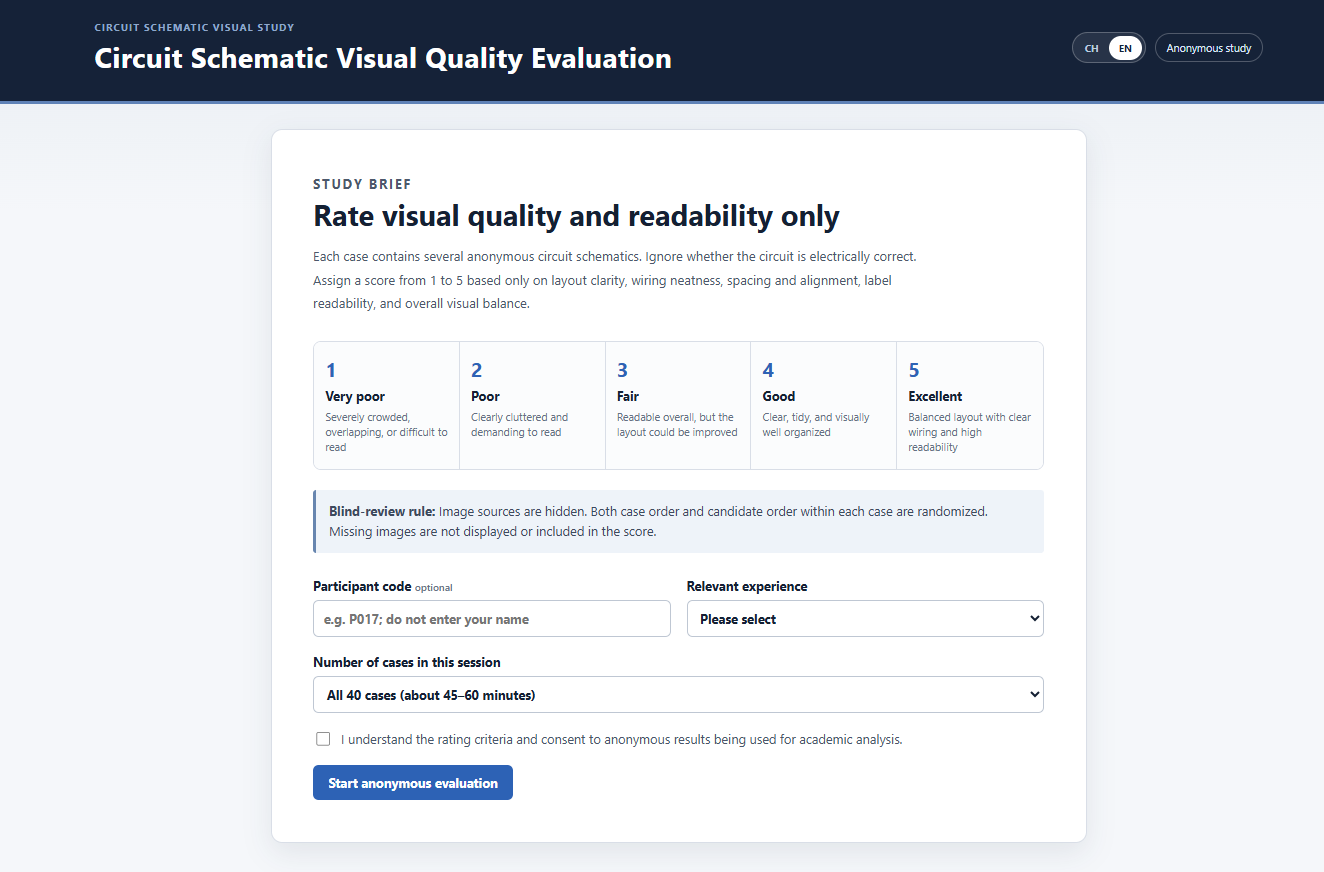}

  \end{minipage}}
  \hfill
  \subfloat[Candidate-scoring interface used during evaluation.]{\label{fig:mos-platform-detail}%
\begin{minipage}[t]{0.48\textwidth}
    \centering
    \includegraphics[
      width=\linewidth,
      height=0.32\textheight,
      keepaspectratio
    ]{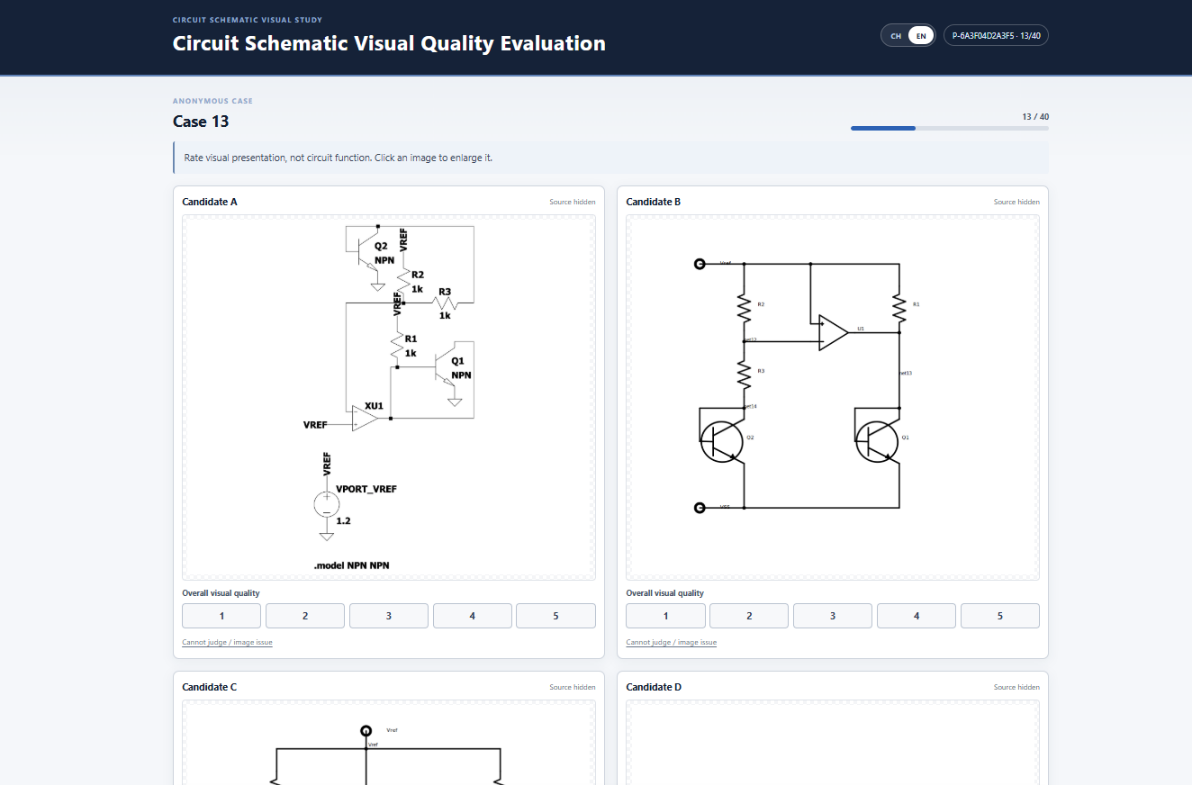}

  \end{minipage}}

  \caption{Blinded MOS evaluation platform: (a) the landing page and
  (b) the candidate-scoring interface.}
  \label{fig:mos-platform}
\end{figure}

\noindent
\begin{minipage}[t]{0.48\textwidth}
\vspace{0pt}
We conducted a blinded mean opinion score (MOS) study on the 40 AMSNet
cases for which reference schematics are available. Fifteen engineers
independently rated the anonymized candidate schematics. The platform
randomized both the case order and the candidate-image order within each
case, and did not reveal the originating method or model.

For each evaluated method, the reported MOS is the arithmetic mean of its
available ratings across participants and cases. Figure~\ref{fig:mos-platform}
shows the study landing page and the rubric presented to participants.

\end{minipage}\hfill
\begin{minipage}[t]{0.48\textwidth}
\vspace{0pt}
Participants were explicitly instructed to ignore electrical functionality
and judge only visual quality and readability, considering layout clarity,
wire cleanliness, spacing and alignment, text legibility, and overall visual
balance. The five-point rubric was:
\begin{enumerate}
  \item \textbf{Very Poor:} severe crowding or overlap makes the schematic
  difficult to read.
  \item \textbf{Poor:} obvious visual clutter imposes a high reading burden.
  \item \textbf{Fair:} the schematic is basically readable, but its
  layout can be improved.
  \item \textbf{Good:} the schematic is clear and orderly, with good visual
  organization.
  \item \textbf{Excellent:} the layout is balanced, connections are clean,
  and the schematic is easy to read.
\end{enumerate}
\end{minipage}

\medskip

\clearpage

\section{Qualitative Comparison with EESchematic and Weave}
\label{app:qualitative-baselines}

\begin{figure}[H]
\centering

\begin{minipage}[t]{0.48\textwidth}
\vspace{0pt}
\centering

\subfloat[Comparator: \method{}.]{\label{fig:comparator-amschem}%
\begin{minipage}[c][0.18\textheight][c]{\linewidth}
    \centering
    \includegraphics[
        width=\linewidth,
        height=0.18\textheight,
        keepaspectratio
    ]{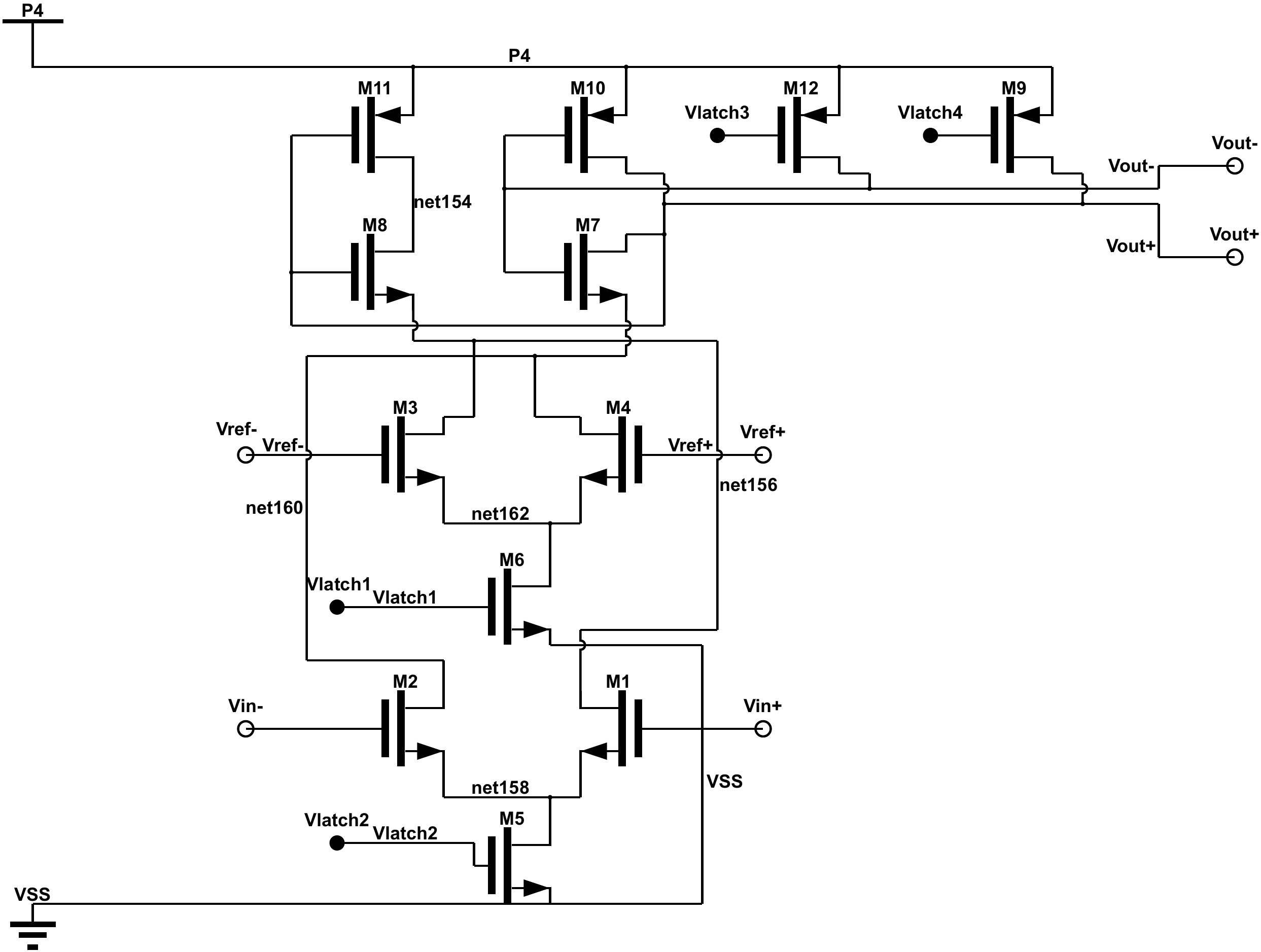}

\end{minipage}}

\vspace{0.6em}

\subfloat[Comparator: EESchematic.]{\label{fig:comparator-eeschematic}%
\begin{minipage}[c][0.15\textheight][c]{\linewidth}
    \centering
    \includegraphics[
        width=\linewidth,
        height=0.15\textheight,
        keepaspectratio
    ]{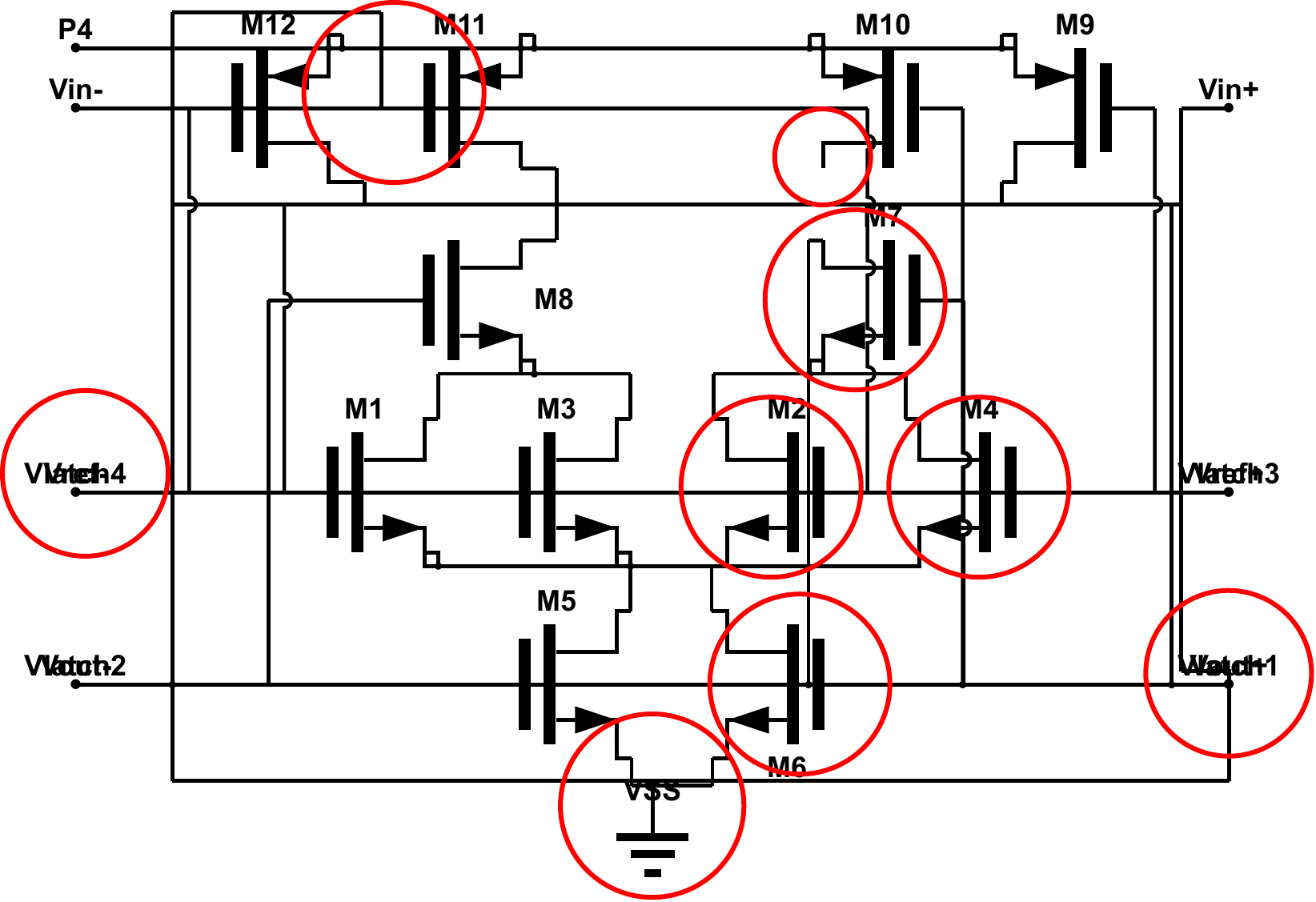}

\end{minipage}}

\vspace{0.6em}

\subfloat[Comparator: Weave.]{\label{fig:comparator-weave}%
\begin{minipage}[c][0.29\textheight][c]{\linewidth}
    \centering
    \includegraphics[
        width=\linewidth,
        height=0.29\textheight,
        keepaspectratio
    ]{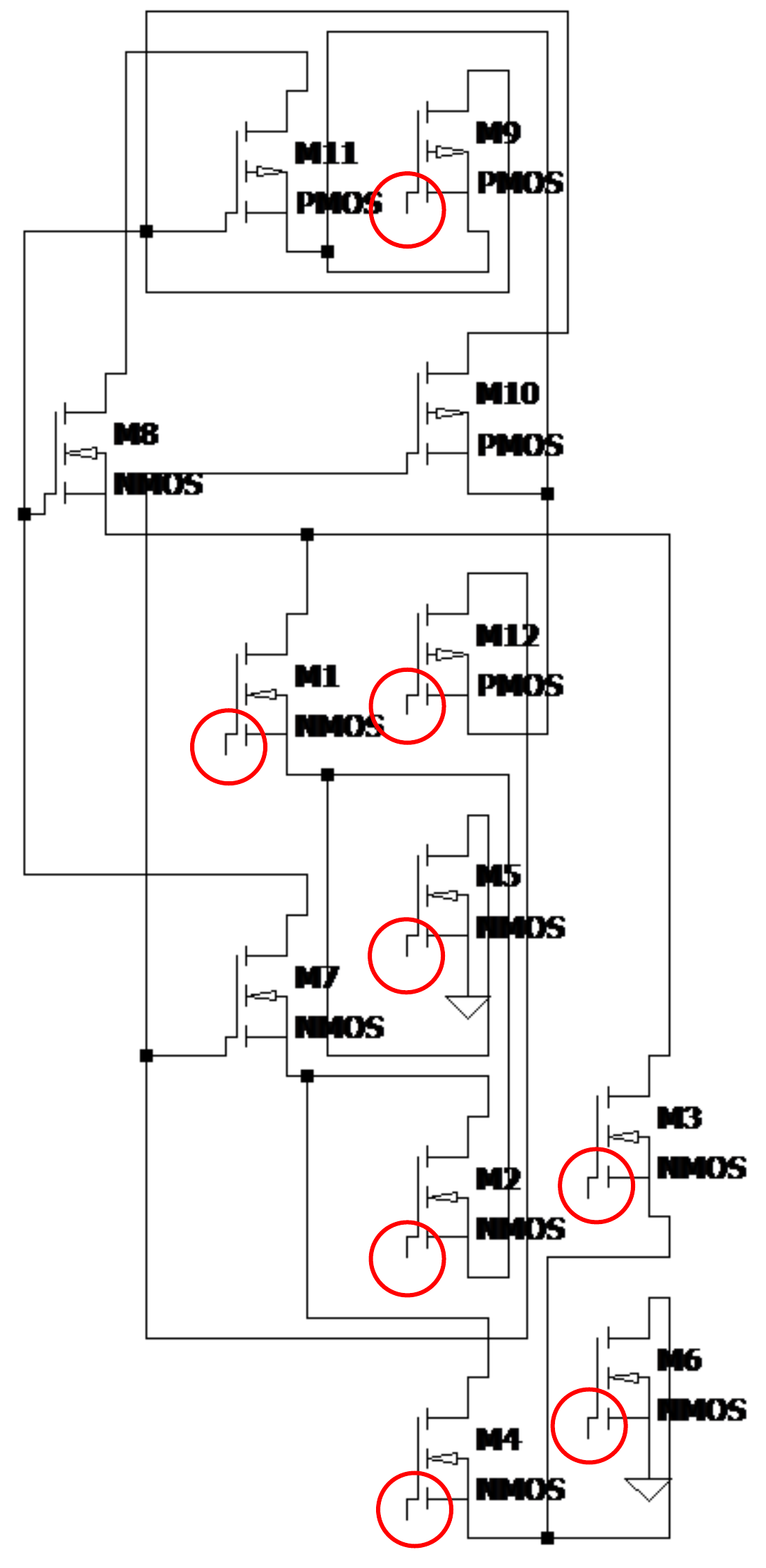}

\end{minipage}}

\end{minipage}
\hfill
\begin{minipage}[t]{0.48\textwidth}
\vspace{0pt}
\centering

\subfloat[Bandgap: \method{}.]{\label{fig:bandgap-amschem}%
\begin{minipage}[c][0.18\textheight][c]{\linewidth}
    \centering
    \includegraphics[
        width=\linewidth,
        height=0.18\textheight,
        keepaspectratio
    ]{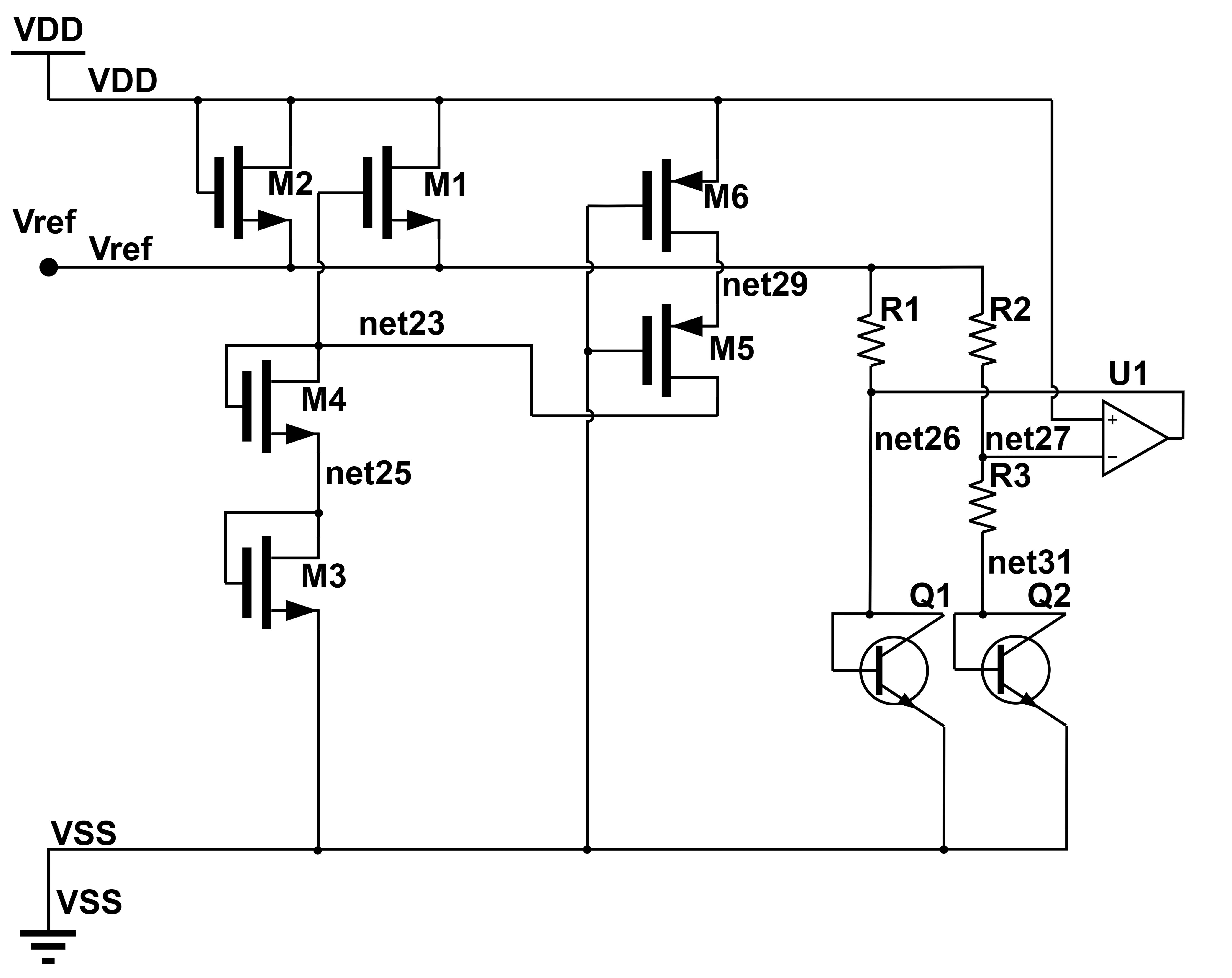}

\end{minipage}}

\vspace{0.6em}

\subfloat[Bandgap: EESchematic.]{\label{fig:bandgap-eeschematic}%
\begin{minipage}[c][0.15\textheight][c]{\linewidth}
    \centering
    \includegraphics[
        width=\linewidth,
        height=0.15\textheight,
        keepaspectratio
    ]{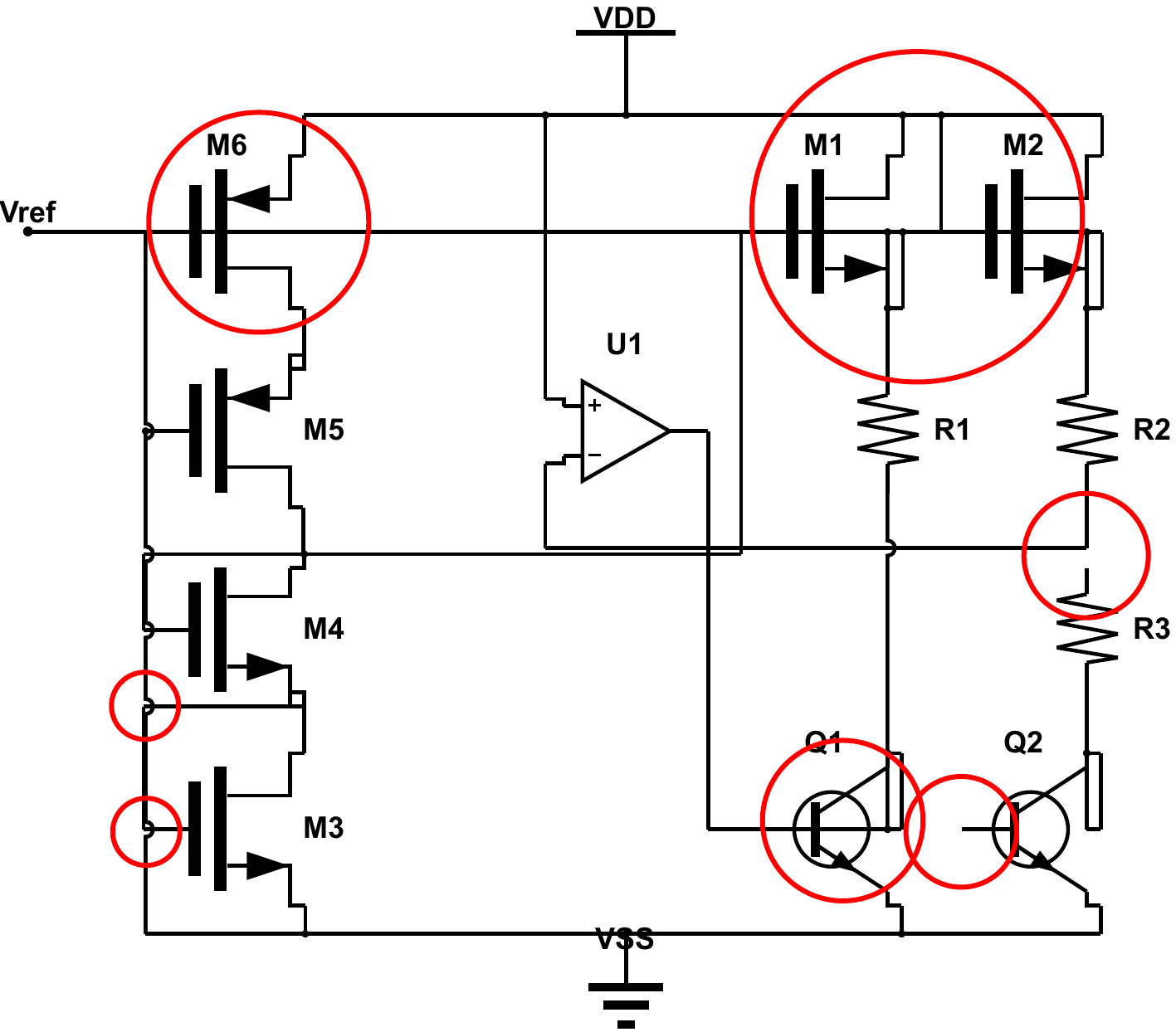}

\end{minipage}}

\vspace{0.6em}

\subfloat[Bandgap: Weave.]{\label{fig:bandgap-weave}%
\begin{minipage}[c][0.29\textheight][c]{\linewidth}
    \centering
    \includegraphics[
        width=\linewidth,
        height=0.29\textheight,
        keepaspectratio
    ]{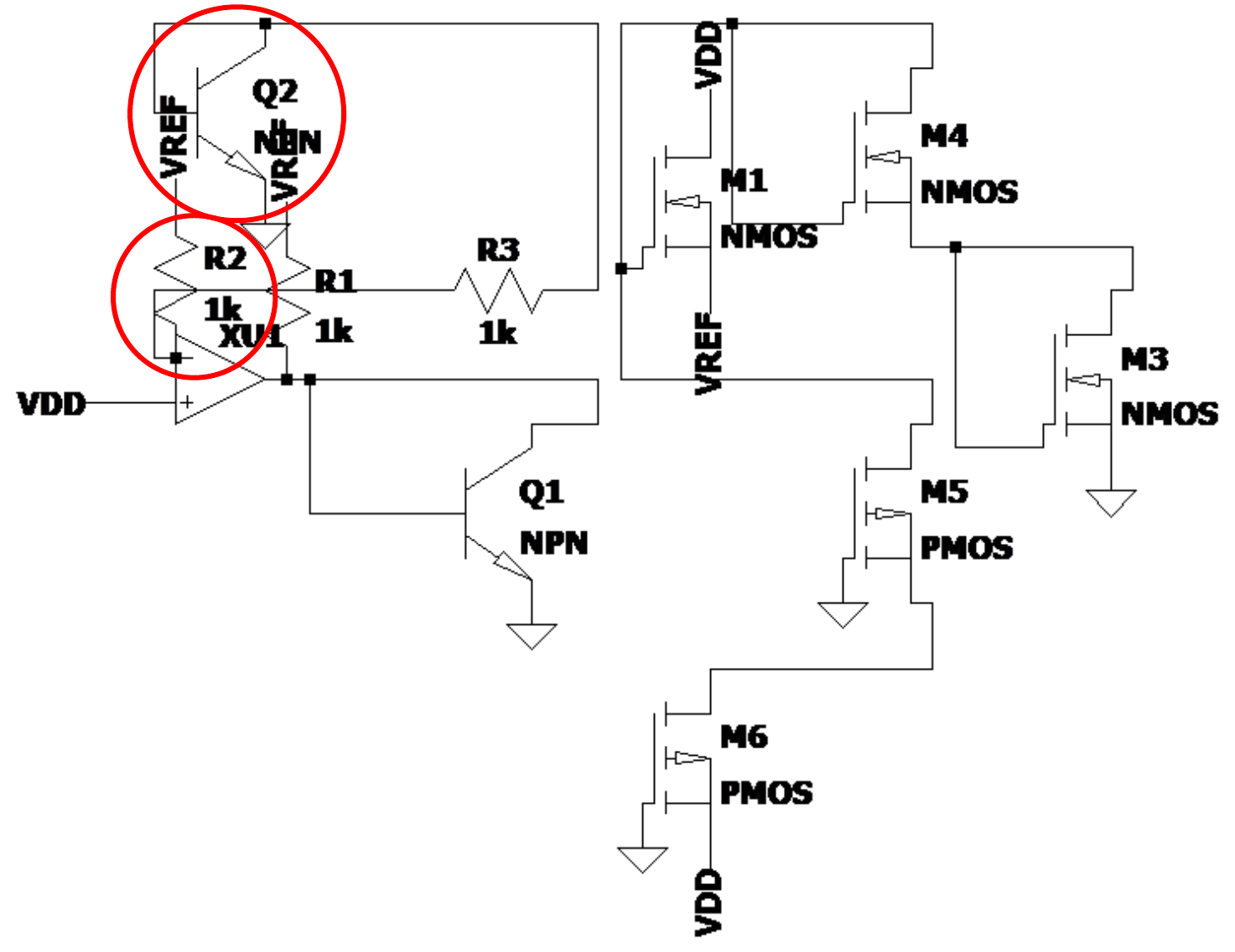}

\end{minipage}}

\end{minipage}

\caption{Qualitative comparison of \method{}, EESchematic, and Weave
on two AMSNet circuits. The left column shows the comparator, and the
right column shows the bandgap reference circuit. Within each column,
the results are ordered from top to bottom as \method{}, EESchematic,
and Weave. All methods use the same input netlist and rendering
settings for each circuit. Red circles in the EESchematic and Weave
outputs mark selected topological and visual errors identified during
manual inspection.}
\label{fig:qualitative-baselines}
\end{figure}

We compare \method{} with our reproduction of
EESchematic~\cite{liu2025eeschematic} and
Weave~\cite{gulgonul2026weave} on the same input netlists.
Figure~\ref{fig:qualitative-baselines} presents the generated
schematics for a comparator and a bandgap circuit.
Selected topological and visual errors in the EESchematic and Weave
outputs are manually highlighted with red circles.

These examples provide a qualitative comparison under the unified
rendering protocol. The red-circle annotations were added by us to
indicate representative errors and are not intended to provide an
exhaustive enumeration of all issues. The EESchematic outputs are
obtained from our reproduction and should not be interpreted as
results from the authors' original implementation.

